\documentclass[10pt,twocolumn,letterpaper]{article}

\usepackage{cvpr}
\usepackage{times}
\usepackage{epsfig}
\usepackage{graphicx}
\usepackage{amsmath}
\usepackage{amssymb}
\usepackage{caption}

\usepackage[pagebackref=true,breaklinks=true,letterpaper=true,colorlinks,bookmarks=false]{hyperref}

\cvprfinalcopy 

\def\cvprPaperID{1950} 

\ifcvprfinal\fi
\begin{document}


\title{DeepEdge: A Multi-Scale Bifurcated Deep Network\\for Top-Down Contour Detection}

\author{Gedas Bertasius\\
University of Pennsylvania\\
{\tt\small gberta@seas.upenn.edu}
\and
Jianbo Shi\\
University of Pennsylvania\\
{\tt\small jshi@seas.upenn.edu}
\and
Lorenzo Torresani\\
Dartmouth College\\
{\tt\small lt@dartmouth.edu}
}

\maketitle

\def\GB#1{{{{\color{magenta} \it #1}}}}
\def\GBR#1{{{{\color{red} REMOVED: \it #1}}}}
\def\LTC#1{{{{\color{green} COMMENT: \it #1}}}}
\def\LT#1{{{{\color{blue} \it #1}}}}

\begin{abstract}

Contour detection has been a fundamental component in many image segmentation and object detection systems. Most previous work utilizes low-level features such as texture or saliency to detect contours and then use them as cues for a higher-level task such as object detection. However, we claim that recognizing objects and predicting contours are two mutually related tasks. Contrary to traditional approaches, we show that we can invert the commonly established pipeline: instead of detecting contours with low-level cues for a higher-level recognition task, we exploit object-related features as high-level cues for contour detection. 

We achieve this goal by means of a multi-scale deep network that consists of five convolutional layers and a bifurcated fully-connected sub-network. The section from the input layer to the fifth convolutional layer is fixed and directly lifted from a pre-trained network optimized over a large-scale object classification task. This section of the network is applied to four different scales of the image input. These four parallel and identical streams are then attached to a bifurcated sub-network consisting of two independently-trained branches. One branch learns to predict the contour likelihood (with a classification objective) whereas the other branch is trained to learn the fraction of human labelers agreeing about the contour presence at a given point (with a regression criterion). 

We show that without any feature engineering our multi-scale deep learning approach achieves state-of-the-art results in contour detection.


\end{abstract}

\section{Introduction}


 \begin{figure}
\begin{center}
   \includegraphics[width=1\linewidth]{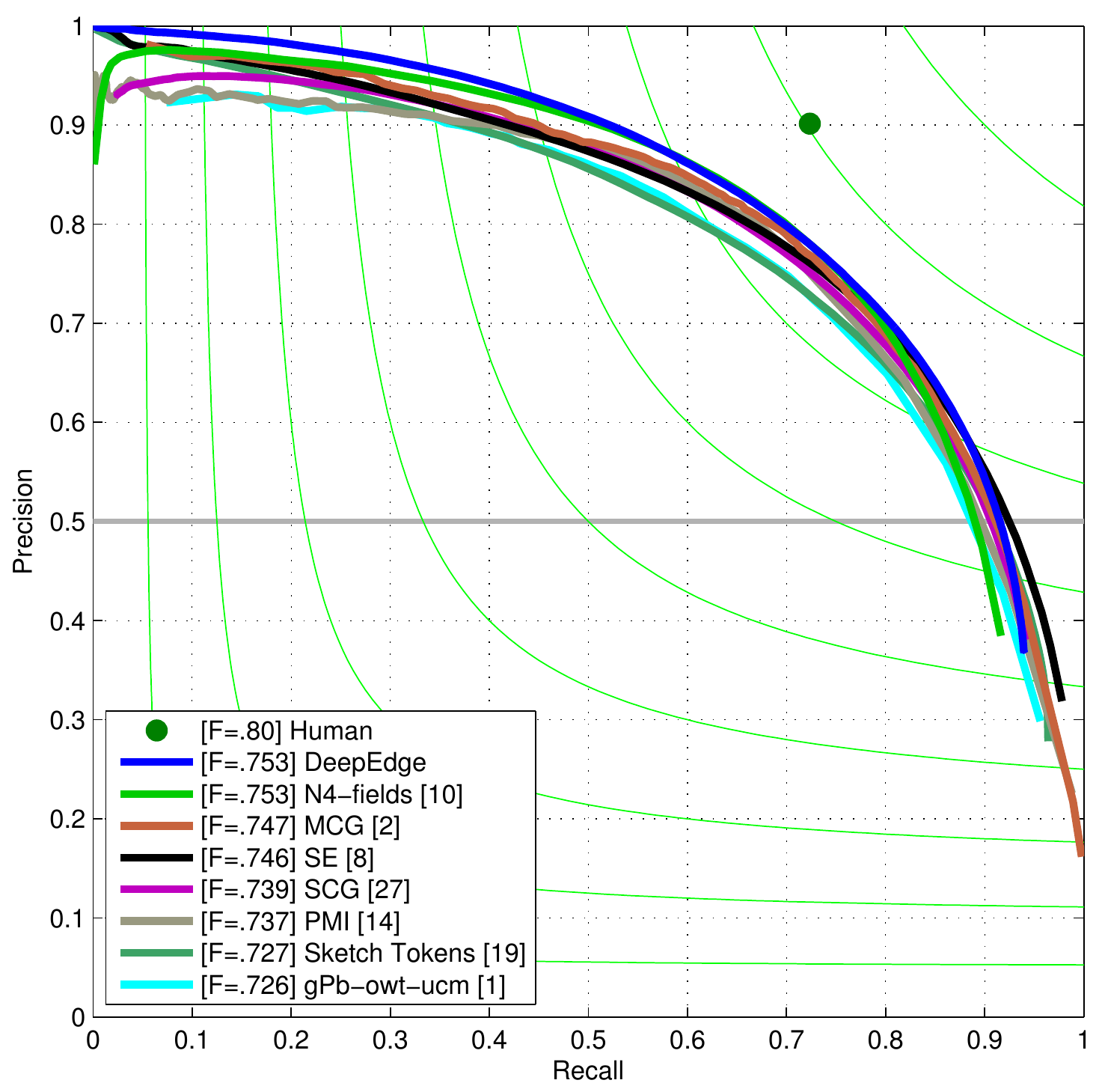}
\end{center}
   \caption{Contour detection accuracy on the BSDS500 dataset. Our method attains higher average precision compared to prior methods and state-of-the-art F-score. At low recall, DeepEdge achieves nearly $100\% $ precision..\vspace{-0.0cm}}
\label{fig:isoF}
\end{figure}


Contour detection is typically considered a low-level problem, and used to aid higher-level tasks such as object detection~\cite{Arbelaez:2011:CDH:1963053.1963088,Borenstein04combiningtop-down,Shotton05contour-basedlearning,Opelt06aboundary-fragment-model}. However, it can be argued that the tasks of detecting objects and predicting contours are closely related. For instance, given the contours we can easily infer which objects are present in the image. Conversely, if we are given exact locations of the objects we could predict contours just as easily.  A commonly established pipeline in computer vision starts with with low-level contour prediction and then moves up to higher-level object detection. However, since we claim that these two tasks are mutually related, we propose to invert this process. Instead of using contours as low-level cues for object detection, we want to use object-specific information as high-level cues for contour detection. Thus, in a sense our scheme can be viewed as a top-down approach where object-level cues inform the low-level contour detection process.




In this work, we present a unified multi-scale deep learning approach that uses higher-level object information to predict contours. Specifically, we present a front-to-end convolutional architecture where contours are learned directly from raw pixels. Our proposed deep learning architecture reuses features computed by the first five convolutional layers of the network of Krizhevsky et al.~\cite{NIPS2012_4824}. We refer to this network as the {\em KNet}. Because the KNet has been trained for object classification, reusing its features enable our method to incorporate object-level information for contour prediction. In the experimental section, we will show that this high-level object information greatly improves contour detection results.




Furthermore, our defined architecture operates on multiple scales simultaneously and combines local and global information from the image, which leads to significantly improved contour detection accuracy rates. 

We connect the features computed by the convolutional layers of the KNet at four different scales of the input with a {\em learned} subnetwork that bifurcates into two branches (the architecture of our model is illustrated in Fig.~\ref{fig:ms_arch}). 

What should the learning objective be?  When a human observer decides if a pixel is a boundary edge, a number of supporting evidence is used with object level reasoning. While it is impossible to record such information, we do have the fraction of observers in agreement for each pixel.  We argue that a learning objective that predicts the fraction of human labelers in agreement can mimic human reasoning better. 

Thus, in the bifurcated sub-network we optimize the two branches with different learning objectives. The weights in one branch are optimized with an edge classification objective, while the other branch is trained to predict the fraction of human labelers in agreement, i.e., using a regression criterion. We show that predictions from the classification branch yield high edge recall, while the outputs of the regression branch have high precision. Thus, fusing these two outputs allows us to obtain excellent results with respect to both metrics and produce state-of-the-art F-score and average precision. 

In summary, the use of higher-level object features, independent optimization of edge classification and regression objectives, as well as a unified multi-scale architecture are the key characteristics that allow our method to achieve the state-of-the-art in contour detection (see Fig.~\ref{fig:isoF}).


%
%

\begin{figure}
\begin{center}
 \includegraphics[width=0.95\linewidth]{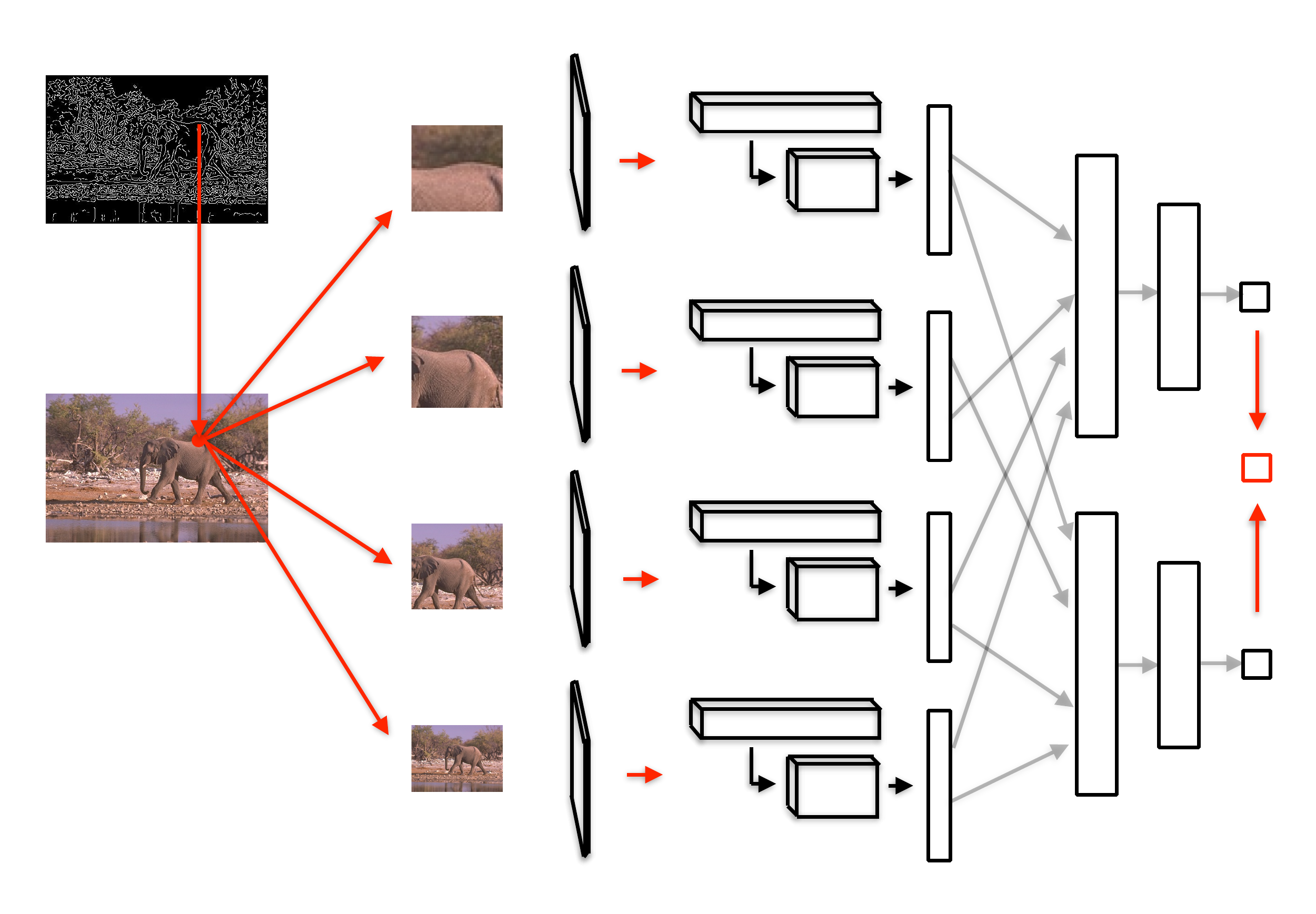}
\end{center}
\vspace{-0.7cm}
   \caption{Visualization of multi-scale DeepEdge network architecture. To extract candidate contour points, we run the Canny edge detector. Then, around each candidate point, we extract patches at four different scales and simultaneously run them through the five convolutional layers of the {\em KNet}~\cite{NIPS2012_4824}. We connect these convolutional layers to two separately-trained network branches. The first branch is trained for classification, while the second branch is trained as a regressor. At testing time, the scalar outputs from these two sub-networks are averaged to produce the final score.\vspace{-0.1cm}}
\label{fig:ms_arch}
\end{figure}


\captionsetup{labelformat=empty}

\begin{figure*}
\centering
\begin{minipage}[b]{.18\textwidth}
\includegraphics[width=1.05\linewidth]{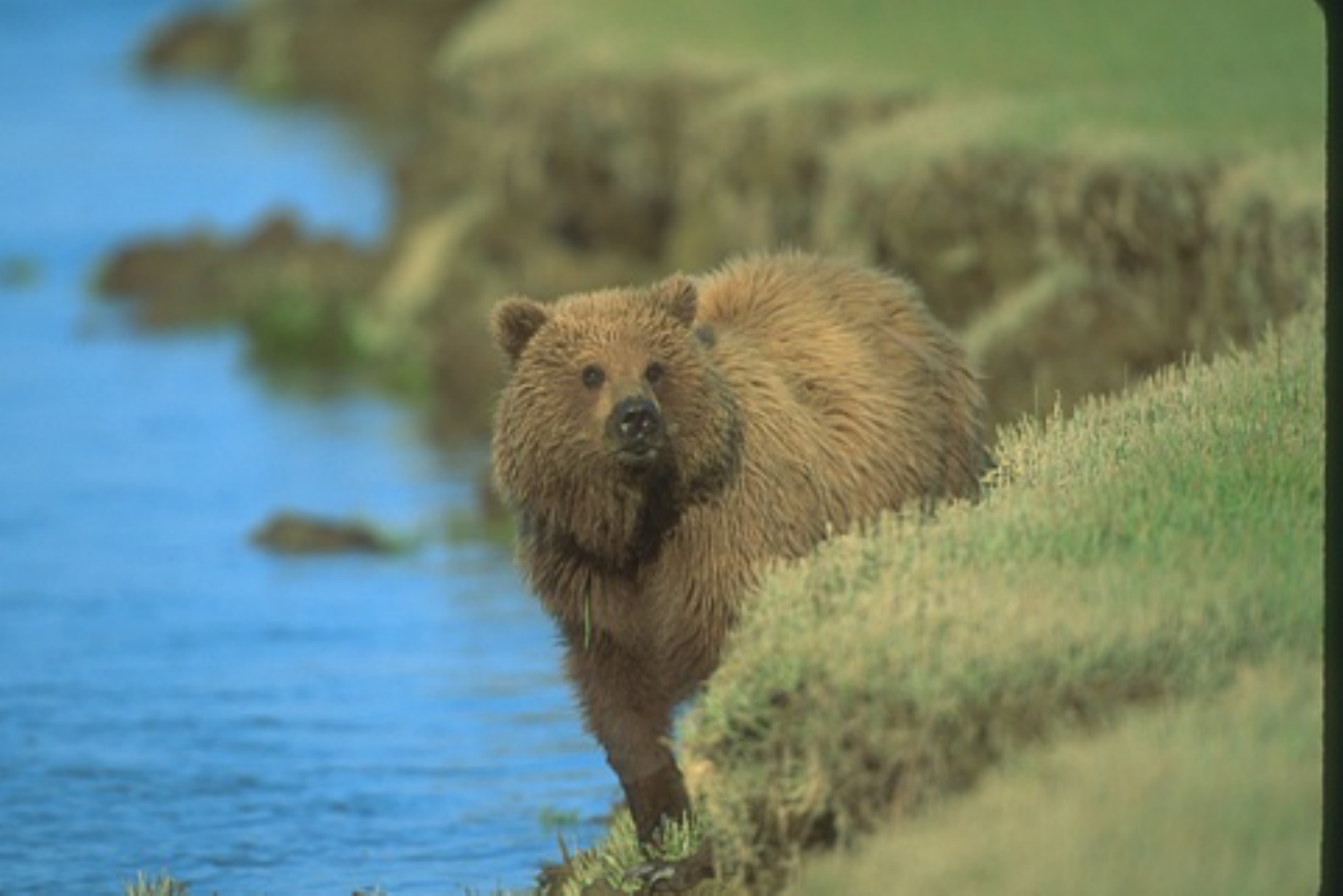}
\caption{a) Input Image}
\end{minipage}
\begin{minipage}[b]{.18\textwidth}
\includegraphics[width=1.05\linewidth]{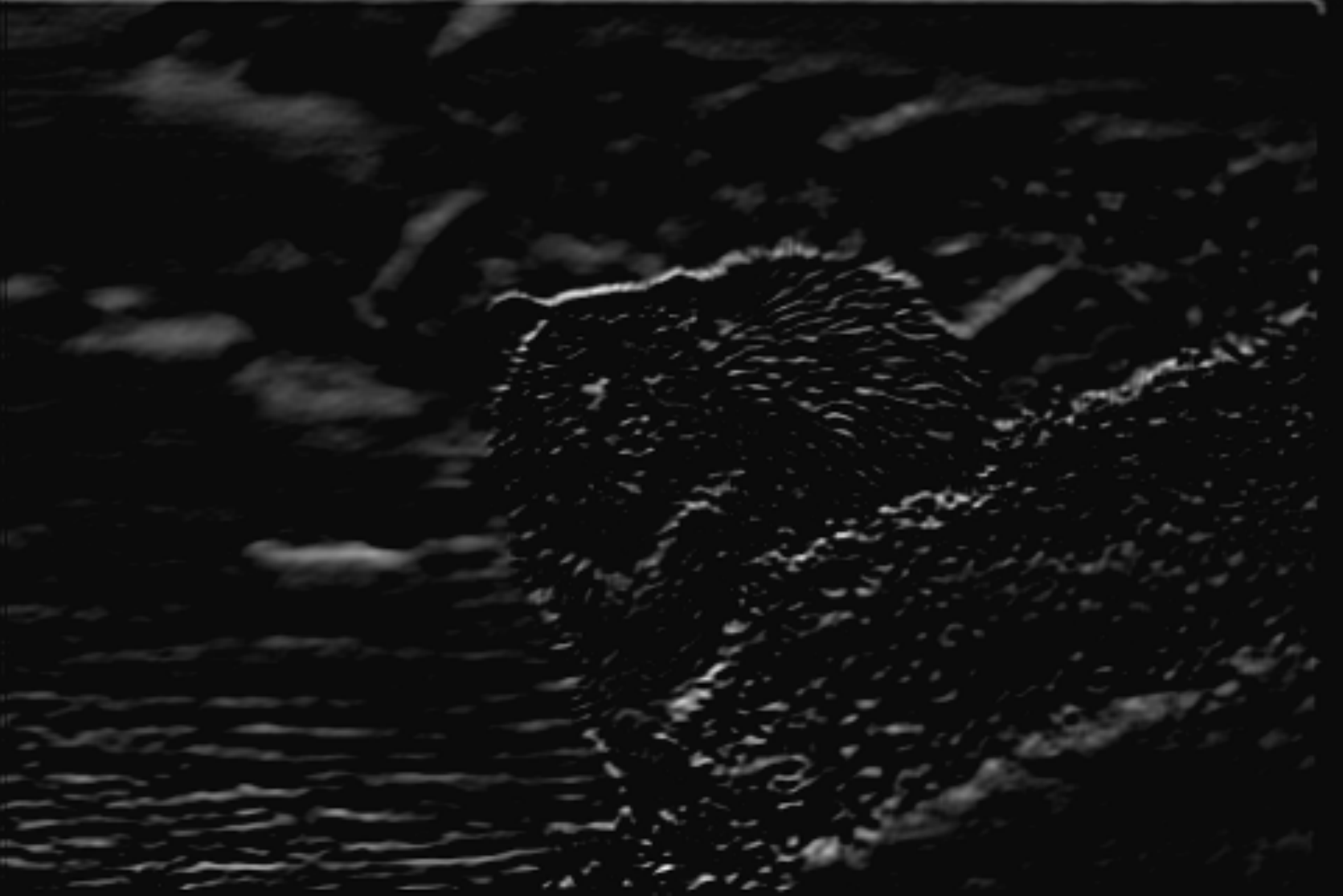}
\caption{c) $2^{nd}$ layer}
\end{minipage}
\begin{minipage}[b]{.18\textwidth}
\includegraphics[width=1.05\linewidth]{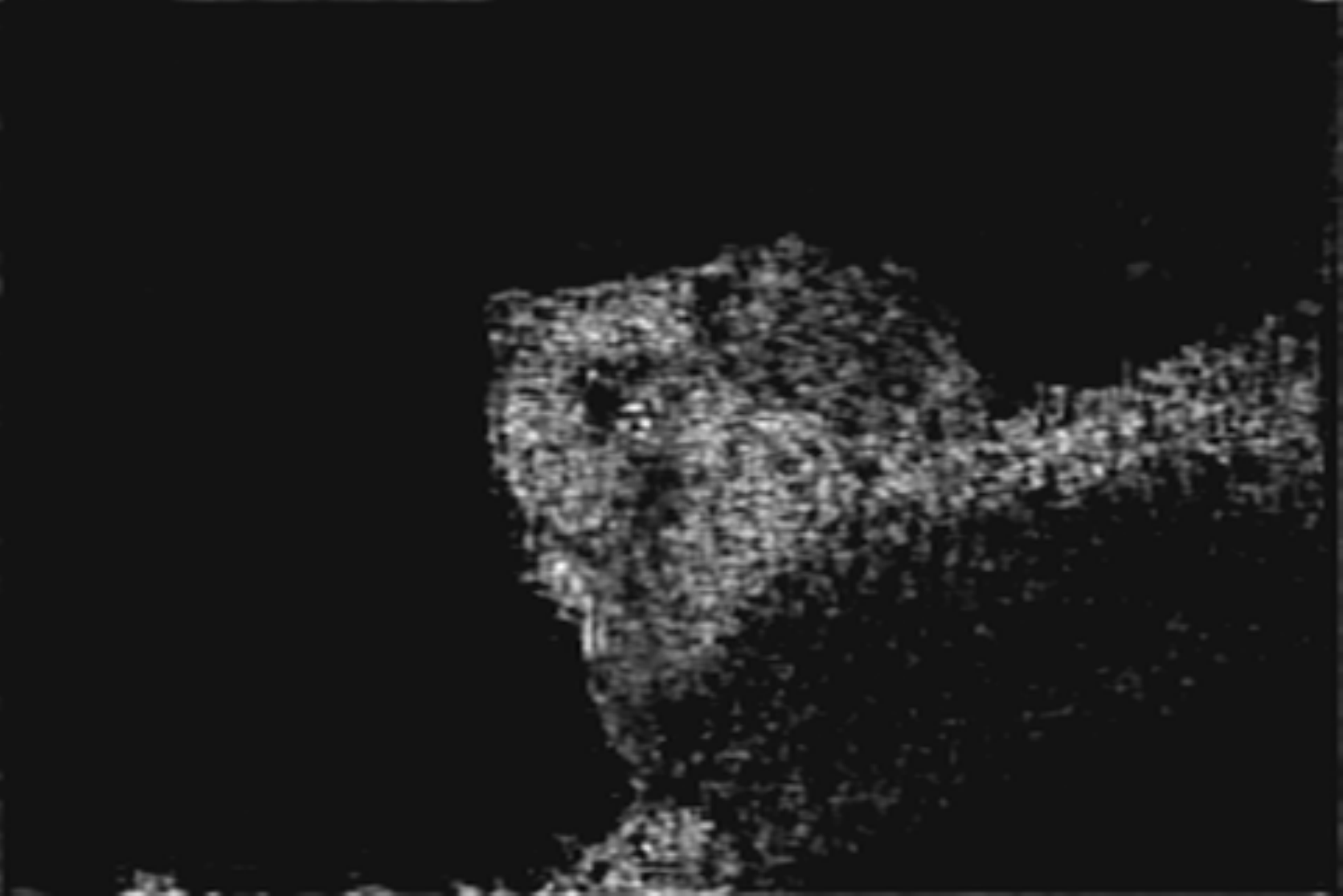}
\caption{d) $3^{rd}$ layer}
\end{minipage}
\begin{minipage}[b]{.18\textwidth}
\includegraphics[width=1.05\linewidth]{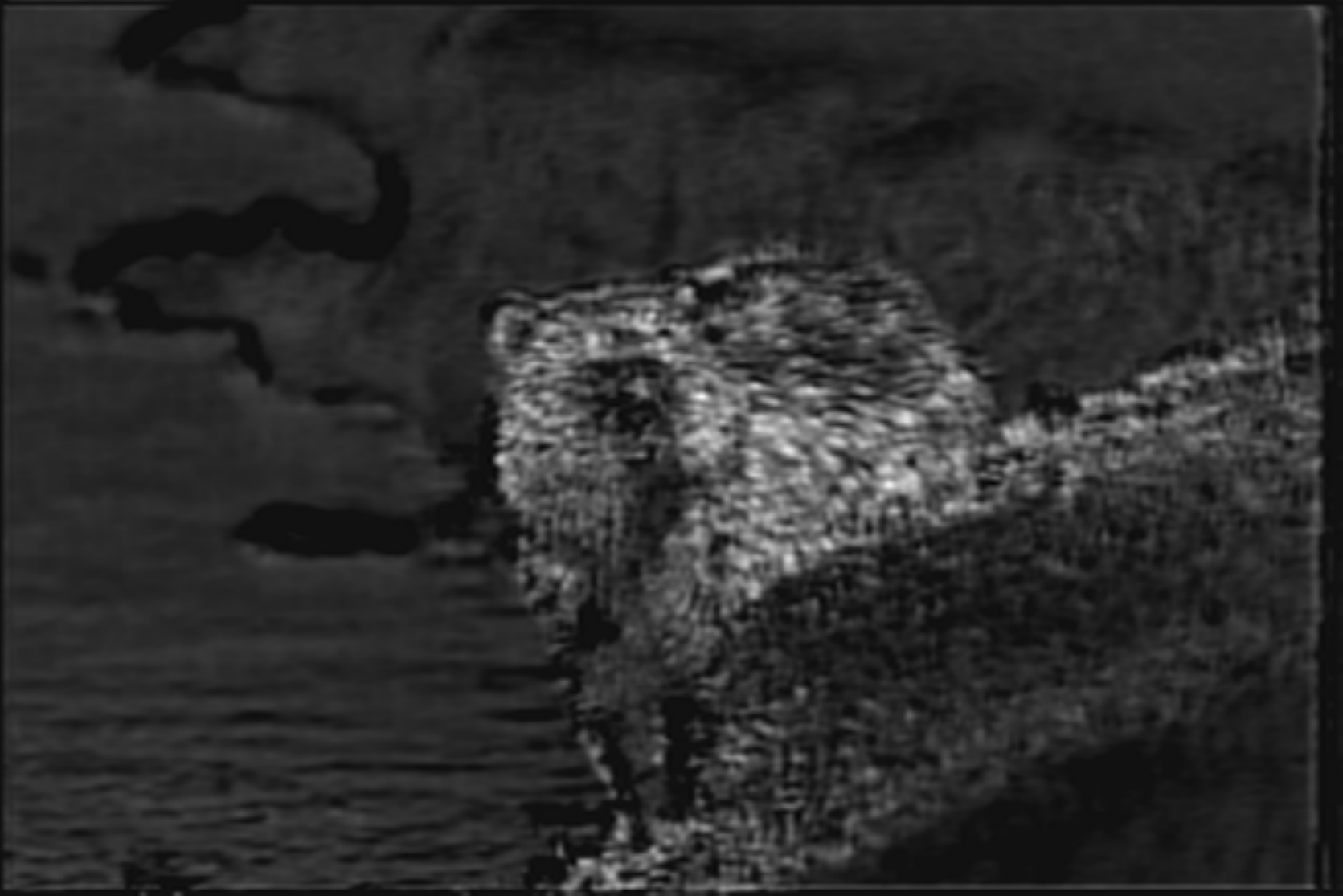}
\caption{e) $4^{th}$ layer}
\end{minipage}
\begin{minipage}[b]{.18\textwidth}
\includegraphics[width=1.05\linewidth]{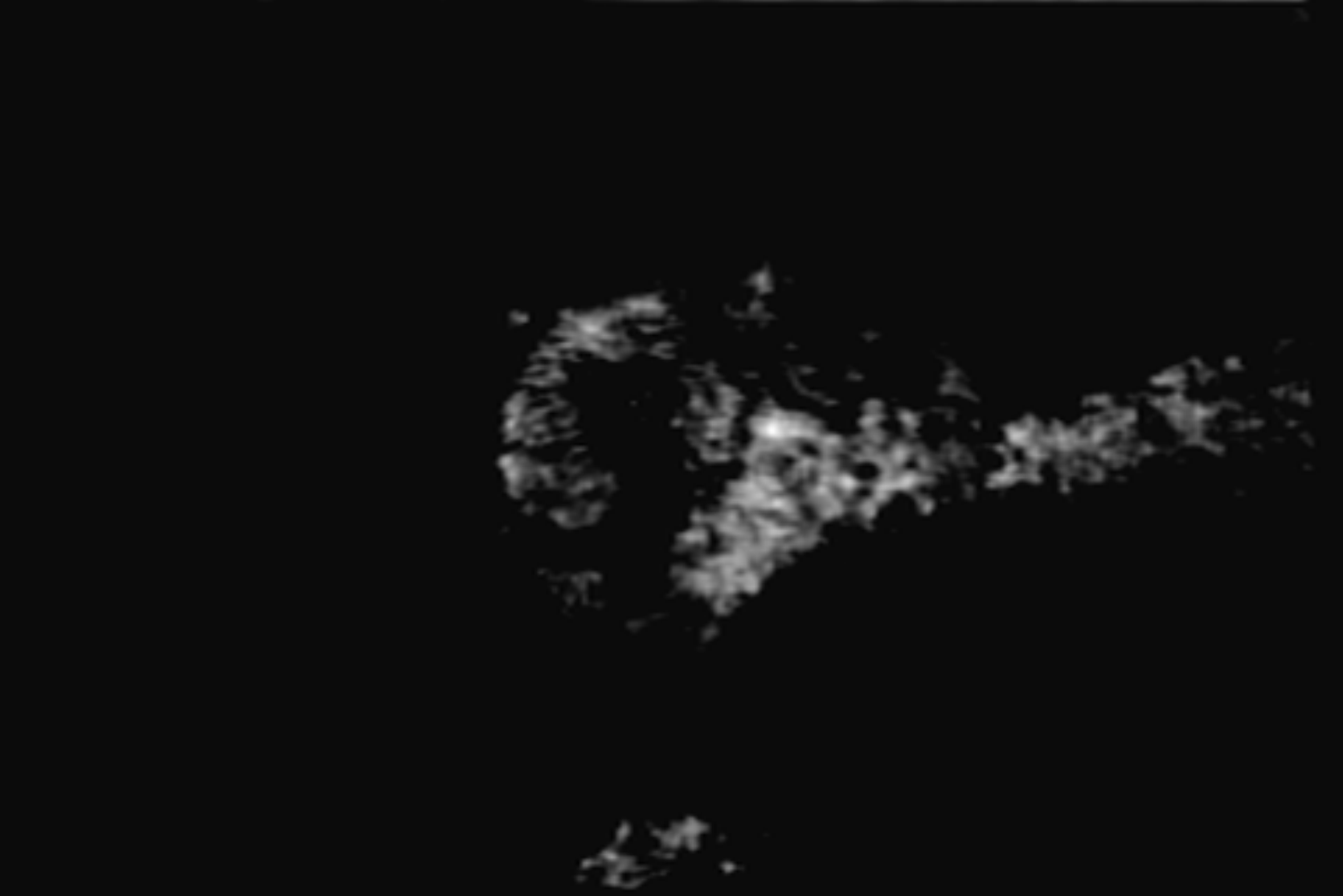}
\caption{e) $5^{th}$ layer}
\end{minipage}
\captionsetup{labelformat=default}
\setcounter{figure}{2}
\caption{Visualization of the activation values at the selected convolutional filters of the {\em KNet} (filters are resized to the original image dimensions). The filters in the second layer fire on oriented edges inside the image. The third and fourth convolutional layers produce an outline of the shape of the object. The fifth layer fires on the specific parts of the object.}
\label{fig:mid_conv}
\end{figure*}

\captionsetup{labelformat=default}

\section{Related Work}

\textbf{Deep Learning.} In the recent years, deep convolutional networks have achieved remarkable results in a wide array of computer vision tasks~\cite{deepFace,DBLP:journals/corr/PinheiroC13,DBLP:journals/corr/ToshevS13,NIPS2012_4824}. However, thus far, applications of convolutional networks focused on high-level vision tasks such as face recognition, image classification, pose estimation or scene labeling~\cite{deepFace,DBLP:journals/corr/PinheiroC13,DBLP:journals/corr/ToshevS13,NIPS2012_4824}. Excellent results in these tasks beg the question whether convolutional networks could perform equally well in lower-level vision tasks such as contour detection. In this paper, we present a convolutional architecture that achieves state-of-the-art results in a contour detection task, thus demonstrating that convolutional networks can be applied successfully for lower-level vision tasks as well.

\textbf{Edge Detection.} Most of the contour detection methods can be divided into two branches: local and global methods. Local methods perform contour detection by reasoning about small patches inside the image. Some recent local methods include sketch tokens~\cite{LimCVPR13SketchTokens} and structured edges~\cite{Dollar2015PAMI}, Both of these methods are trained in a supervised fashion using a random forest classifier. Sketch tokens~\cite{LimCVPR13SketchTokens} pose contour detection as a multi-class classification task and predicts a label for each of the pixels individually. Structured edges~\cite{Dollar2015PAMI}, on the other hand, attempt to predict the labels of multiple pixels simultaneously. 

Global methods predict contours based on the information from the full image. Some of the most successful approaches in this genre are the MCG detector~\cite{cArbelaez14}, gPb detector~\cite{Arbelaez:2011:CDH:1963053.1963088} and sparse code gradients~\cite{ren_nips12}. While sparse code gradients use supervised SVM learning~\cite{Burges98atutorial}, both gPb and MCG rely on some form of spectral methods. Other spectral-based methods include Normalized Cuts~\cite{Shi97normalizedcuts} and PMI~\cite{crisp_boundaries}.

Recently, there have also been attempts to apply deep learning methods to the task of contour detection. While SCT~\cite{MYP:ACCV:2014} is a sparse coding approach, both $N^4$ fields~\cite{DBLP:journals/corr/GaninL14}  and DeepNet~\cite{kivinen2014visual} use Convolutional Neural Networks (CNNs) to predict contours. $N^4$ fields rely on dictionary learning and the use of the Nearest Neighbor algorithm within a CNN framework while DeepNet uses a traditional CNN architecture to predict contours.

In comparison to these prior approaches, our work offers several contributions. First, we define a novel multi-scale bifurcated CNN architecture that enables our network to achieve state-of-the-art contour detection results. Second, we avoid manual feature engineering by learning contours directly from raw data. Finally, we believe  that we are the first to propose the use of high-level object features for contour detection, thus inverting the traditional pipeline relying on low-level cues for mid-level feature extraction. Our experiments show that this top-down approach for contour detection yields state-of-the-art results.


 \captionsetup{labelformat=empty}

\begin{figure*}
\centering
\begin{minipage}[b]{.24\textwidth}
\includegraphics[width=1\linewidth]{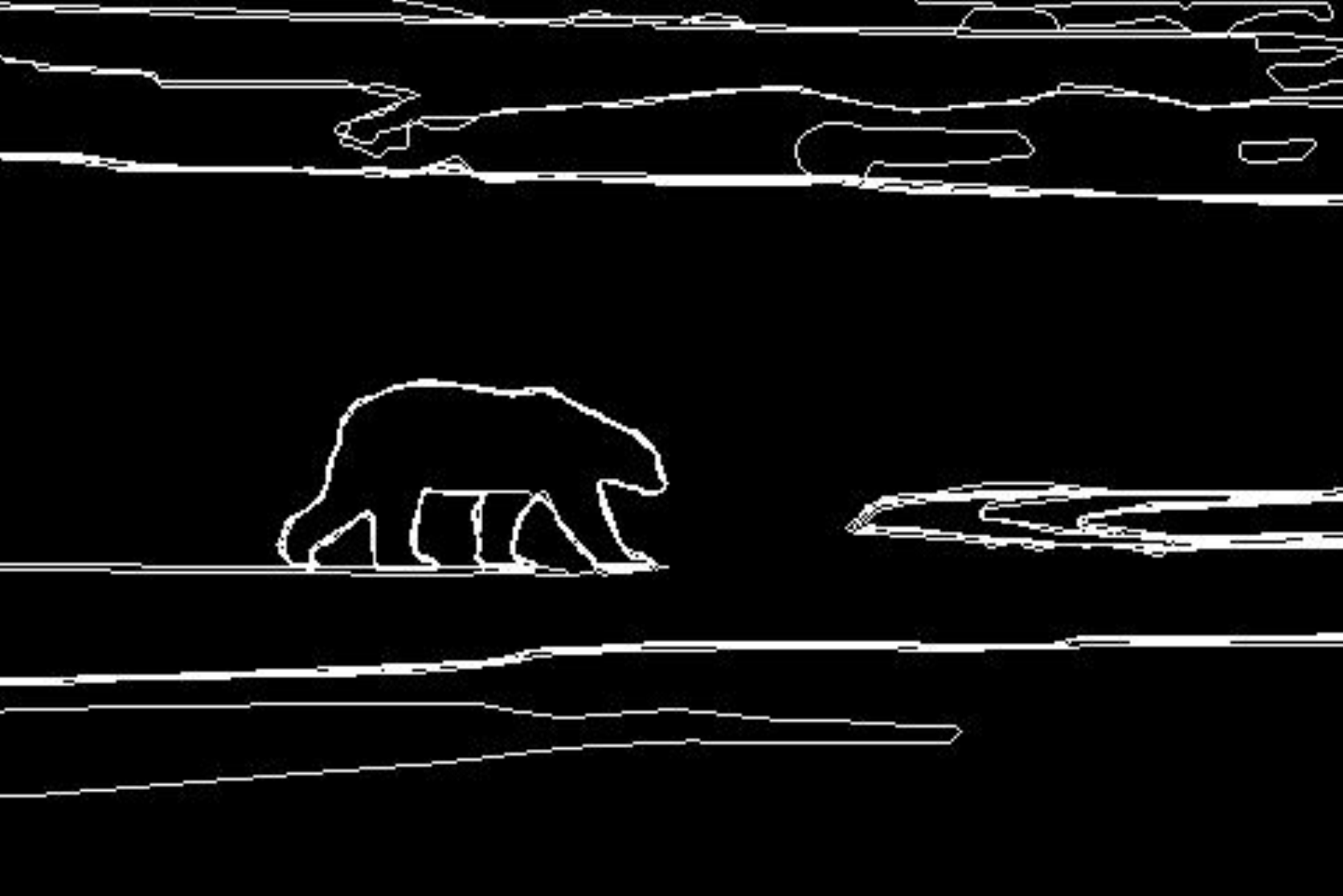}
\end{minipage}
\begin{minipage}[b]{.24\textwidth}
\includegraphics[width=1\linewidth]{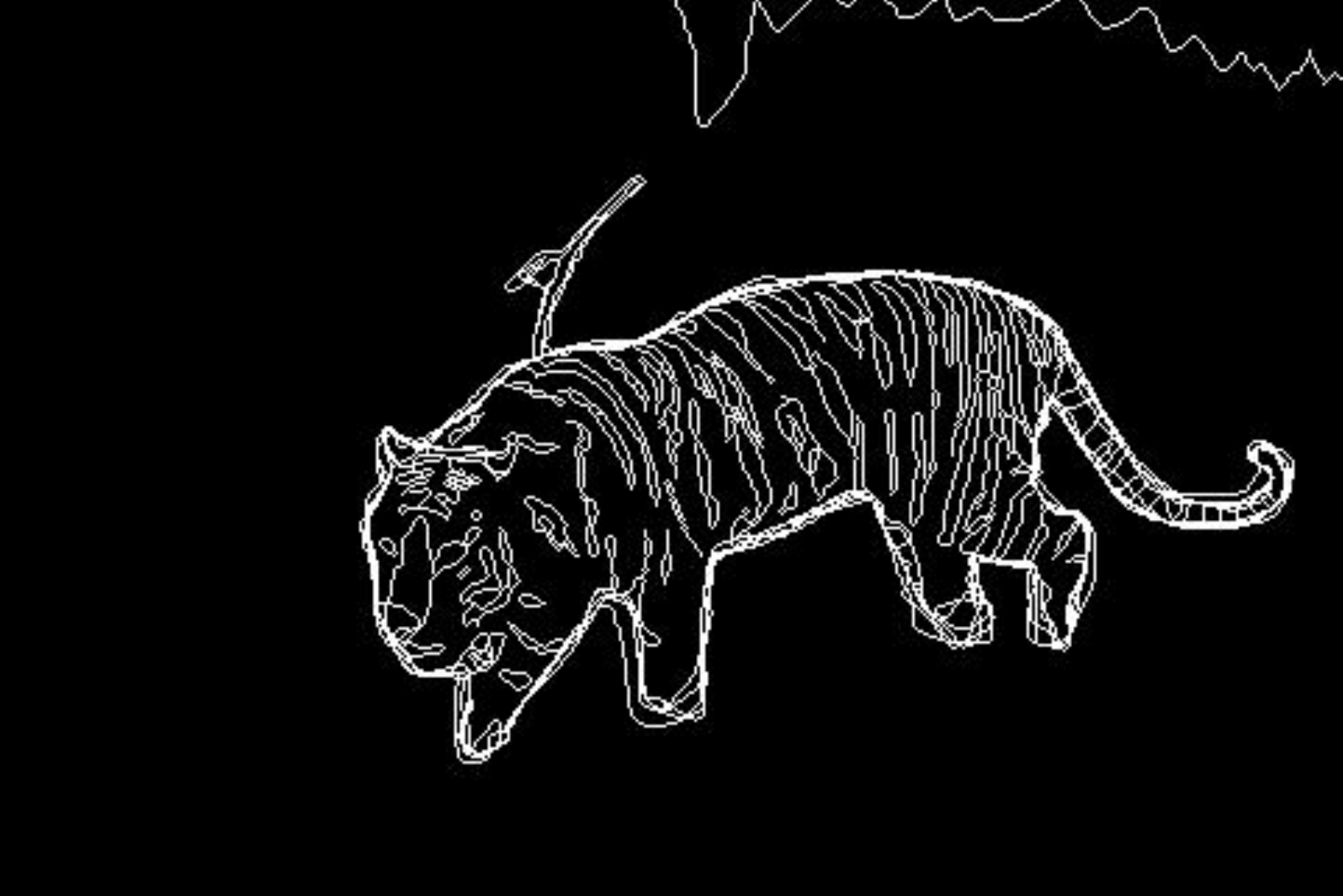}
\end{minipage}
\begin{minipage}[b]{.24\textwidth}
\includegraphics[width=1\linewidth]{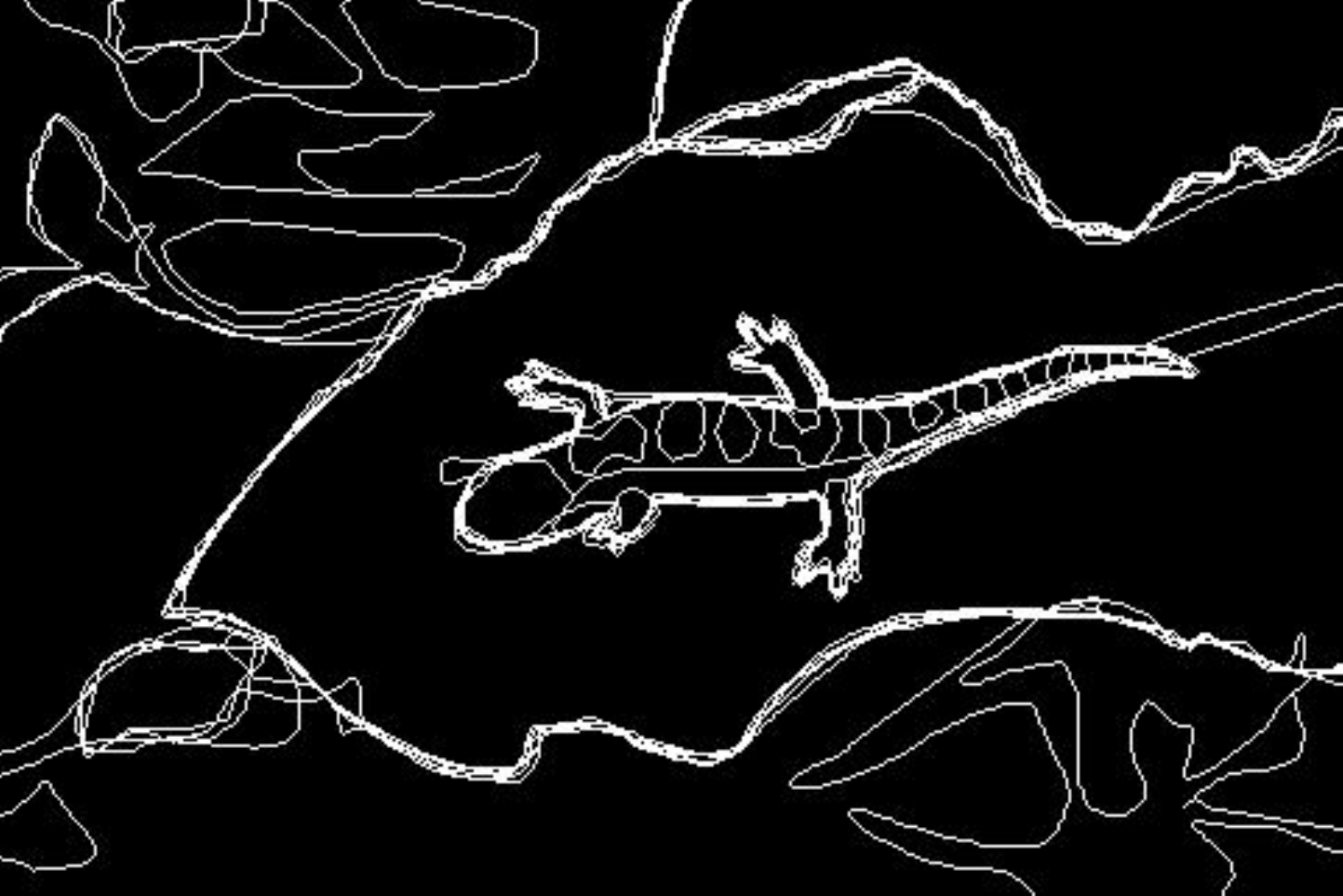}
\end{minipage}
\begin{minipage}[b]{.24\textwidth}
\includegraphics[width=1\linewidth]{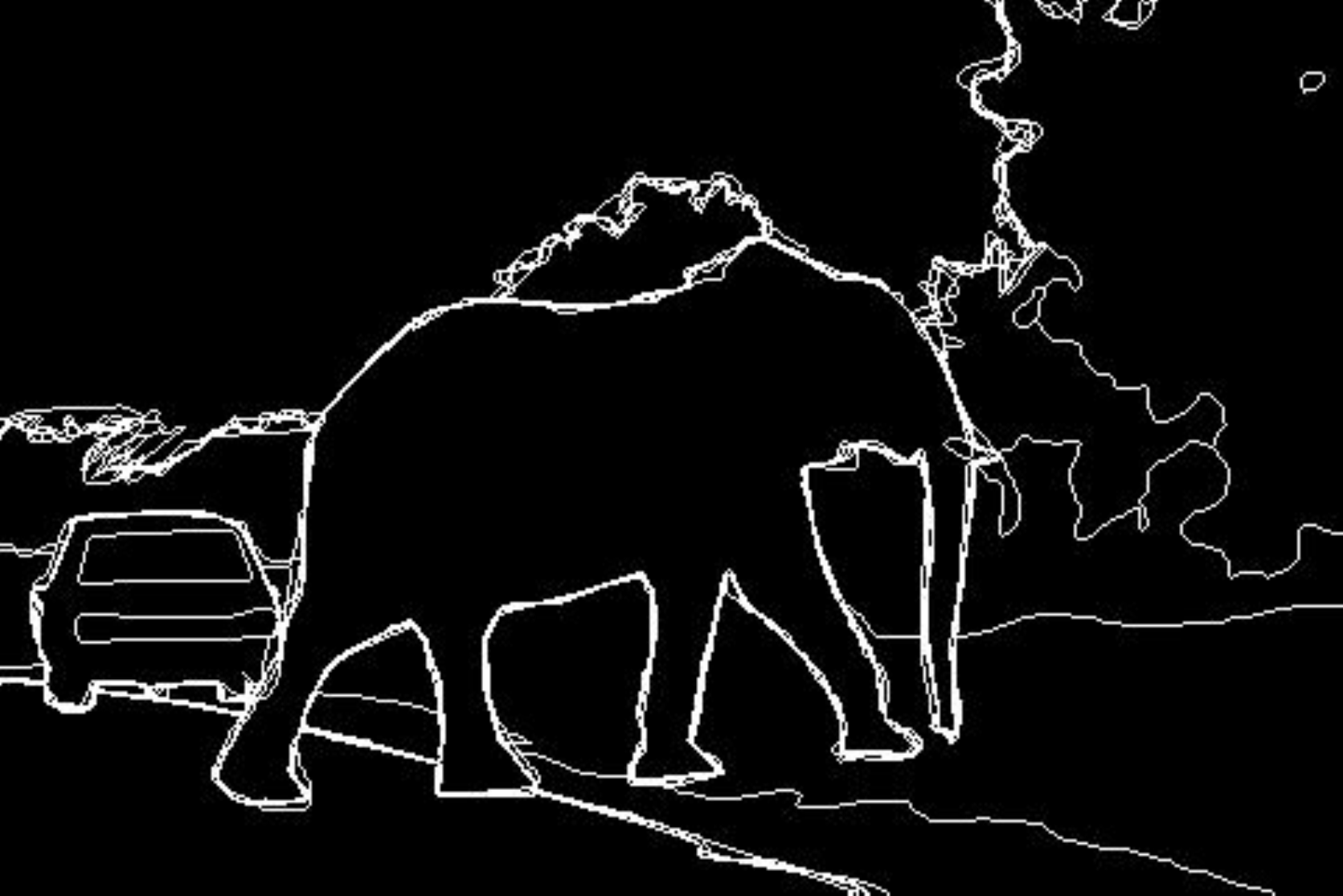}
\end{minipage}

\vspace{0.5mm}

\begin{minipage}[b]{.24\textwidth}
\includegraphics[width=1\linewidth]{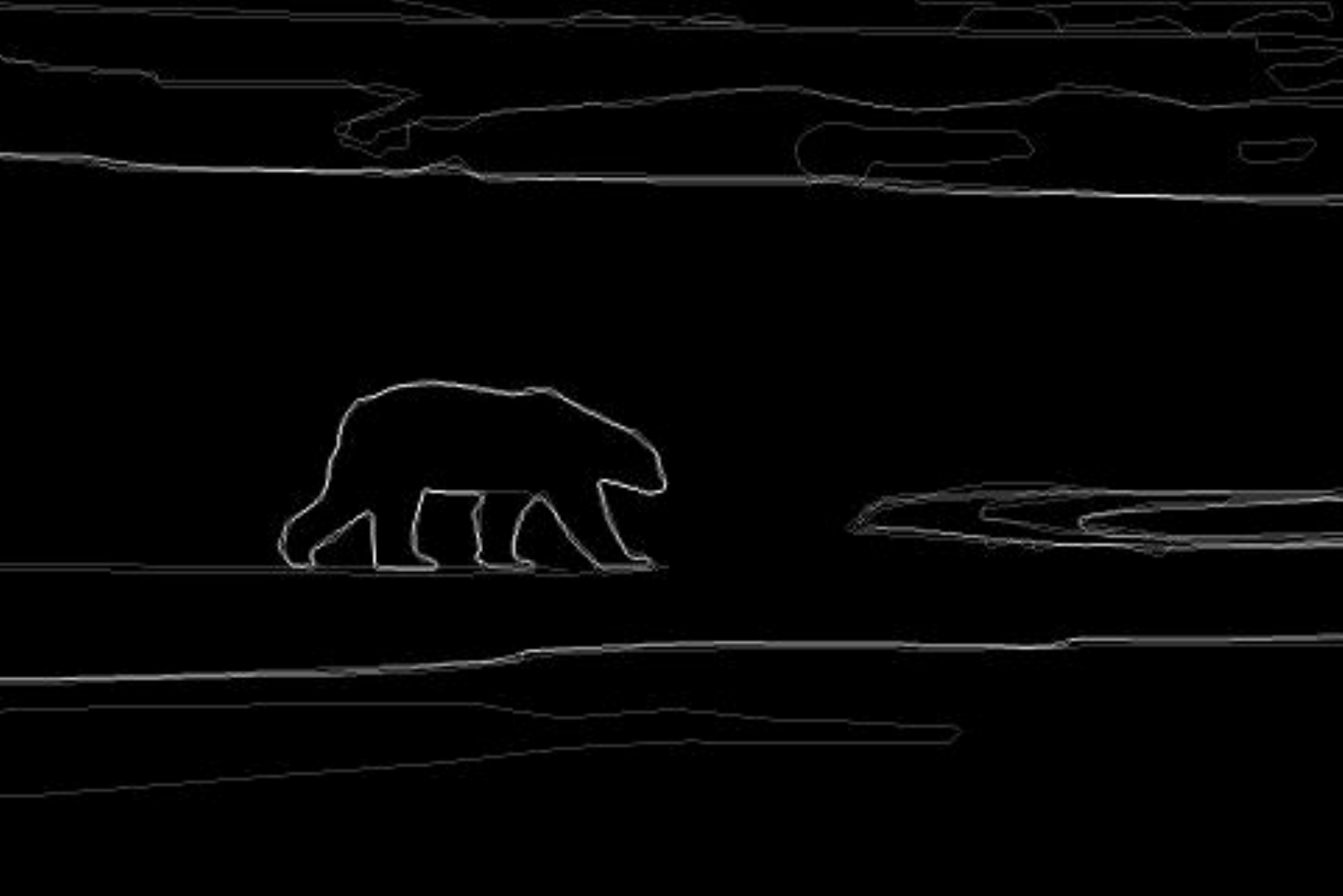}
\end{minipage}
\begin{minipage}[b]{.24\textwidth}
\includegraphics[width=1\linewidth]{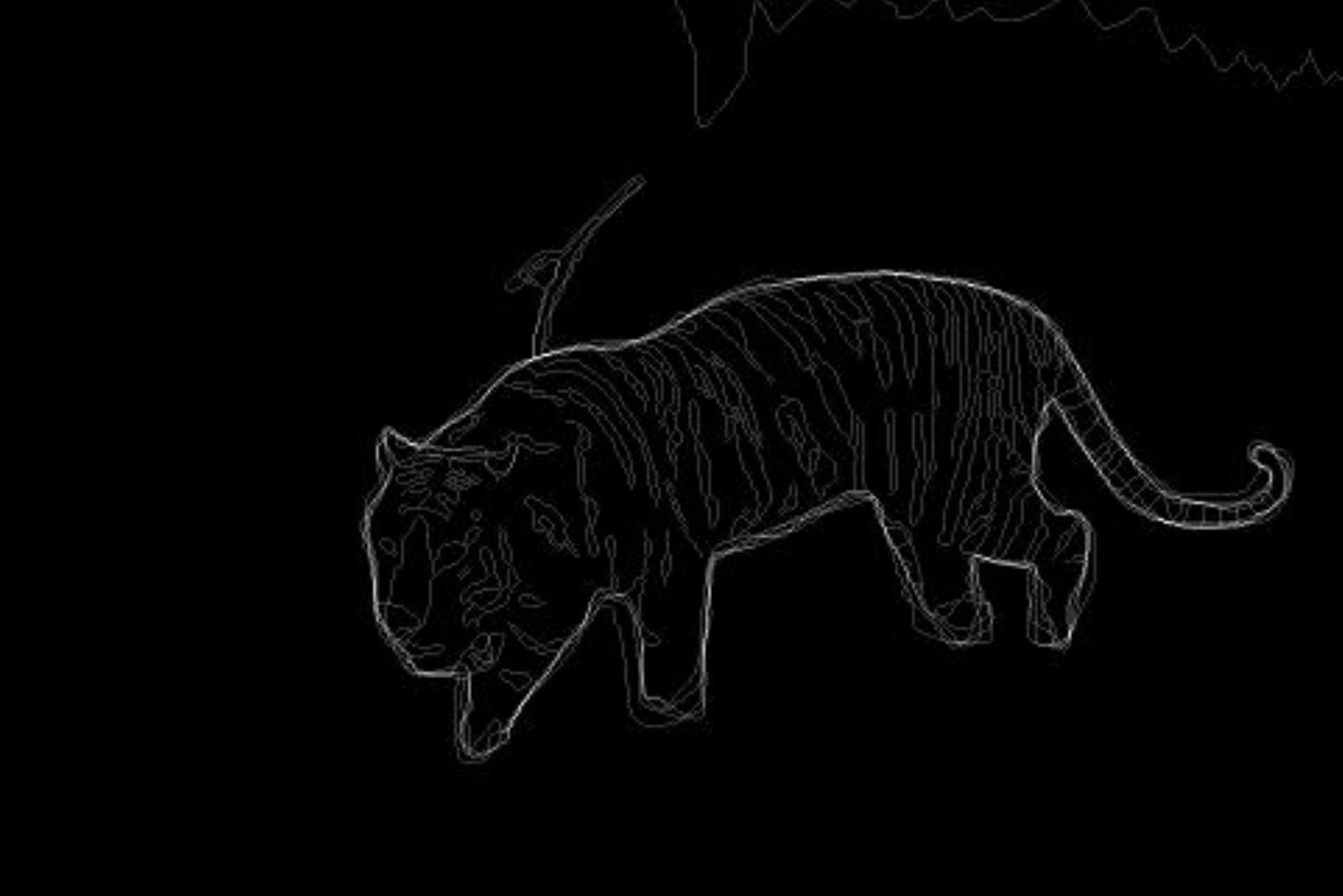}
\end{minipage}
\begin{minipage}[b]{.24\textwidth}
\includegraphics[width=1\linewidth]{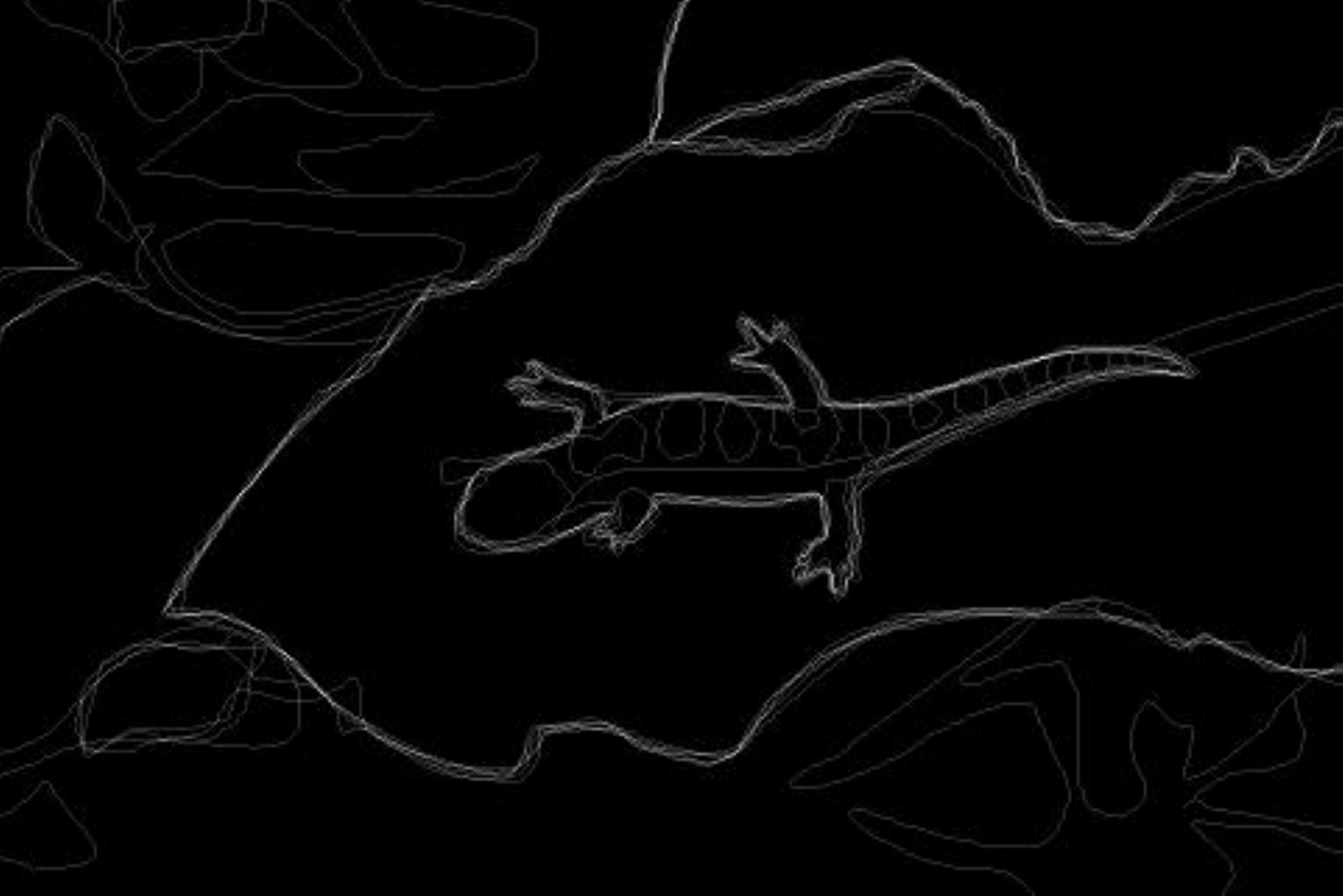}
\end{minipage}
\begin{minipage}[b]{.24\textwidth}
\includegraphics[width=1\linewidth]{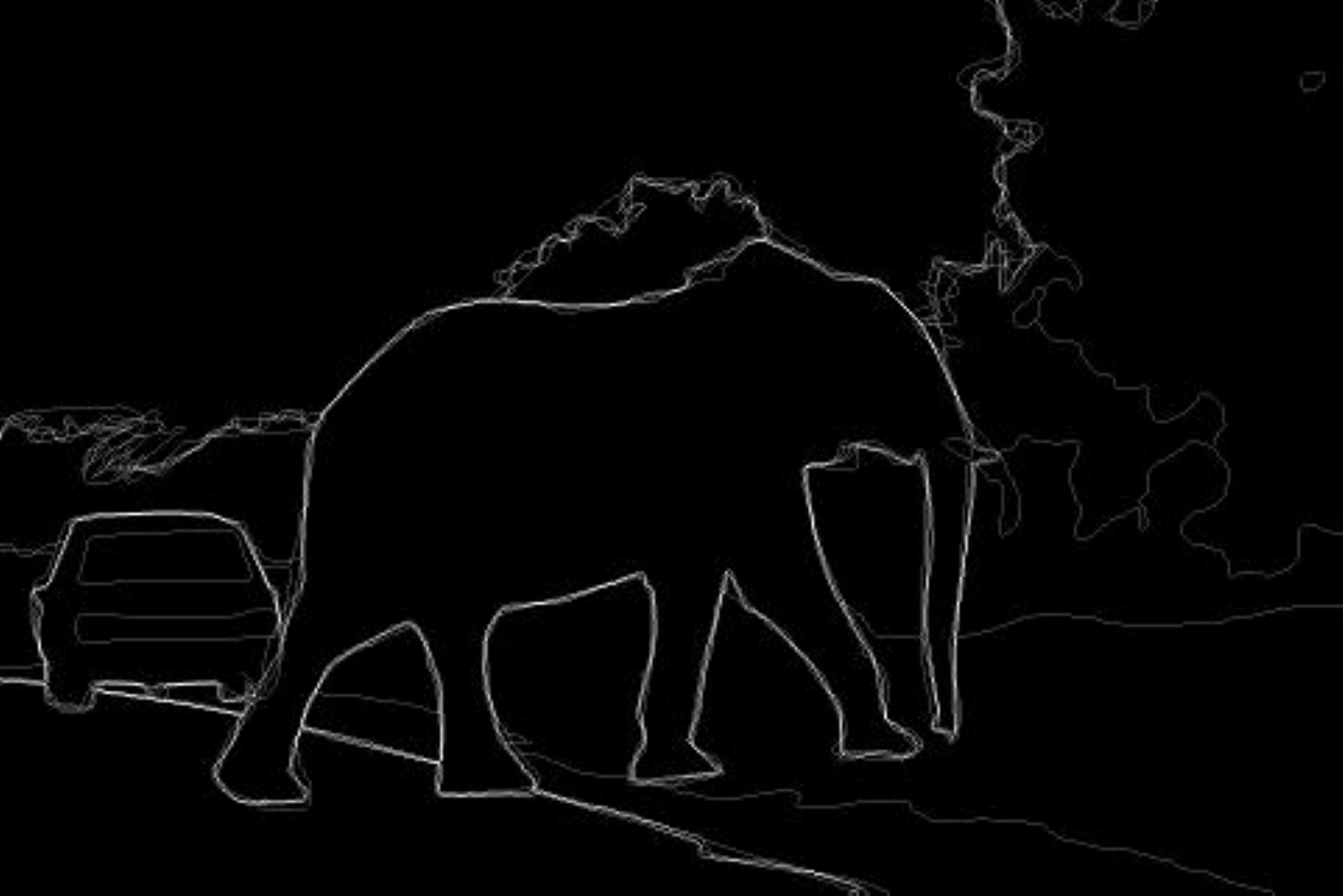}
\end{minipage}

\setcounter{figure}{3}
\captionsetup{labelformat=default}
   \caption{A few samples of ground truth data illustrating the difference between the classification (first row) and the regression (second row) objectives. The classification branch is trained to detect contours that are marked by at least one of the human annotators. Conversely, the regression branch is optimized to the contour values that depict the fraction of human annotators agreeing on the contour.}
\label{fig:reg_vs_class}
\end{figure*}

\captionsetup{labelformat=default}



\section{The DeepEdge Network}


In this section, we describe our proposed deep learning approach for contour detection. 
For simplicity, we first present our architecture in the single-scale scenario (subsection~\ref{sec:single}) and then discuss how to take advantage of multiple scales (subsection~\ref{sec:multiple}).

\subsection{Single-Scale Architecture}
\label{sec:single}

{\bf Selection of Candidate Edge Points.} To extract a set of candidate contours with high recall we apply the Canny edge detector~\cite{Canny:1986:CAE:11274.11275} to the input image. For each of these points we then extract a patch of fixed size  such that our candidate point is the center of the patch. Patches that do not fit into the image boundaries are padded with the mirror reflections of itself.  


{\bf Extraction of High-Level Features.} We then resize the patch of fixed size to match the input dimensions of the KNet~\cite{NIPS2012_4824} and use this network to extract object-level features. The KNet is an appropriate model for our setting as it has been trained over a large number of object classes (the $1000$ categories of the ImageNet dataset~\cite{ILSVRCarxiv14}) and thus captures features that are generic and useful for many object categories. While such features have been optimized for the task of object class recognition, they have been shown to be highly effective for other image-analysis tasks, including object detection~\cite{girshick2014rcnn}, attribute prediction~\cite{journals/corr/ZhangPRDB13}, and image style recognition~\cite{journals/corr/KarayevHWAD13}. The network was trained on $1.2$ million images and it includes more than $60$ million parameters. Its architecture consists of $5$ convolutional layers and $3$ fully connected layers. As we intend to use the KNet as a feature extractor for boundary detection, we utilize only the first $5$ convolutional layers, which preserve explicit location information before the ``spatial scrambling'' of the fully connection layers (note that a spatially variant representation is crucially necessary to predict the presence of contours at individual pixels). The first two KNet convolutional layers learn low-level information. As we move into the deeper layers, however, we observe that the network learns higher-level object information. The second convolutional layer seems to encode coherent edge structures. The third convolutional layer fires at locations corresponding to prototypical object shapes. The fourth layer appears to generate high responses for full shapes of the object, whereas the fifth layer fires on the specific object parts (See Fig.~\ref{fig:mid_conv}). 

In order to obtain a representation that captures 
this hierarchical information, we perform feature extraction at each of the five convolutional layers, as shown in Fig.~\ref{fig:ss_arch}. Specifically, we consider a small sub-volume of the feature map stack produced at each layer. The sub-volume is centered at the center of the patch in order to assess the presence of a contour in a small area around the candidate point. We then perform {\em max}, {\em average}, and {\em center} pooling on this sub-volume. This yields a feature descriptor of size $3 \times F$ where $F$ is the number of feature maps computed by the convolutional layer. While max and average pooling are well established operations in deep learning, we define center pooling as selecting the center-value from each of the feature maps. The motivation for center pooling is that for each candidate point we want to predict the contour presence at that particular point. Because the candidate point is located at the center of the input patch, center pooling extracts the activation value from the location that corresponds to our candidate point location.

{\bf A Bifurcated Sub-Network.}
We connect the feature maps computed via pooling from the five convolutional layers to two separately-trained network branches. Each branch consists of two fully-connected layers. The first branch is trained using binary labels, i.e., to perform contour classification. This branch is making less selective predictions by classifying whether a given point is a contour or not. In a sense, this classification branch abstracts details related to the edge structure (orientation, strength, etc) and simply tries to predict the presence/absence of an edge at a given point. Due to such abstractions, the classification branch produces contour predictions with high recall.






The second branch is optimized as a regressor to predict the fraction of human labelers agreeing about the contour presence at a particular point. Due to a regression objective, this branch is much more selective than the first branch. Intuitively, the second branch is trained to learn the structural differences between the contours that are marked by a different fraction of human labelers. For instance, the area that was labeled as a contour by $80\%$ of human labelers must be significantly different than the area that was labeled as a contour by $20\%$ human labelers. The regression branch is trying to learn such differences by predicting the fraction of human labelers who would mark a particular point as a contour. Thus, in a sense, we are training the regression branch to implicitly mimic how human labelers reason about the contour presence at a given point. In the experimental section, we demonstrate that due to its selectivity, the regression branch produces contour predictions with very high precision. In Fig.~\ref{fig:reg_vs_class}, we present several samples of ground truth data that illustrate the different properties of our two end-objectives.

The number of hidden layers in the first and second fully connected layers of both branches are $1024$ and $512$, respectively. Both branches optimize the sum of squared difference loss over the (binary or continuous) labels. At testing time, the scalar outputs computed from these two sub-networks are averaged to produce a final score indicative of the probability that the candidate point is a contour. Visualization of this architecture is presented in Fig.~\ref{fig:ss_arch}.

In order to train our sub-network, we generate patch examples and labels using training images with ground truth annotations from multiple human labelers. To generate the binary labels, we first sample $40,000$ positive examples that were marked as contours by at least one of the labelers. To generate negative examples we consider the points that were selected as candidate contour points by the Canny edge detector but that have not been marked as contours by any of the human labelers. These are essentially false positives. For training, we use a random subset of $40,000$ of such points in order to have equal proportion of negative and positive examples. These $80,000$ examples are then used to train our classification sub-network. 


In addition to the binary labels, we also generate continuous labels that are used to train the regression network. For this purpose, we define the regression label of a point to be the fraction of human labelers that marked the point as a contour. These $80,000$ examples with continuous labels are then also used to train our regression sub-network.

%

%
%

\begin{figure*}
\begin{center}
 \includegraphics[width=0.85\linewidth]{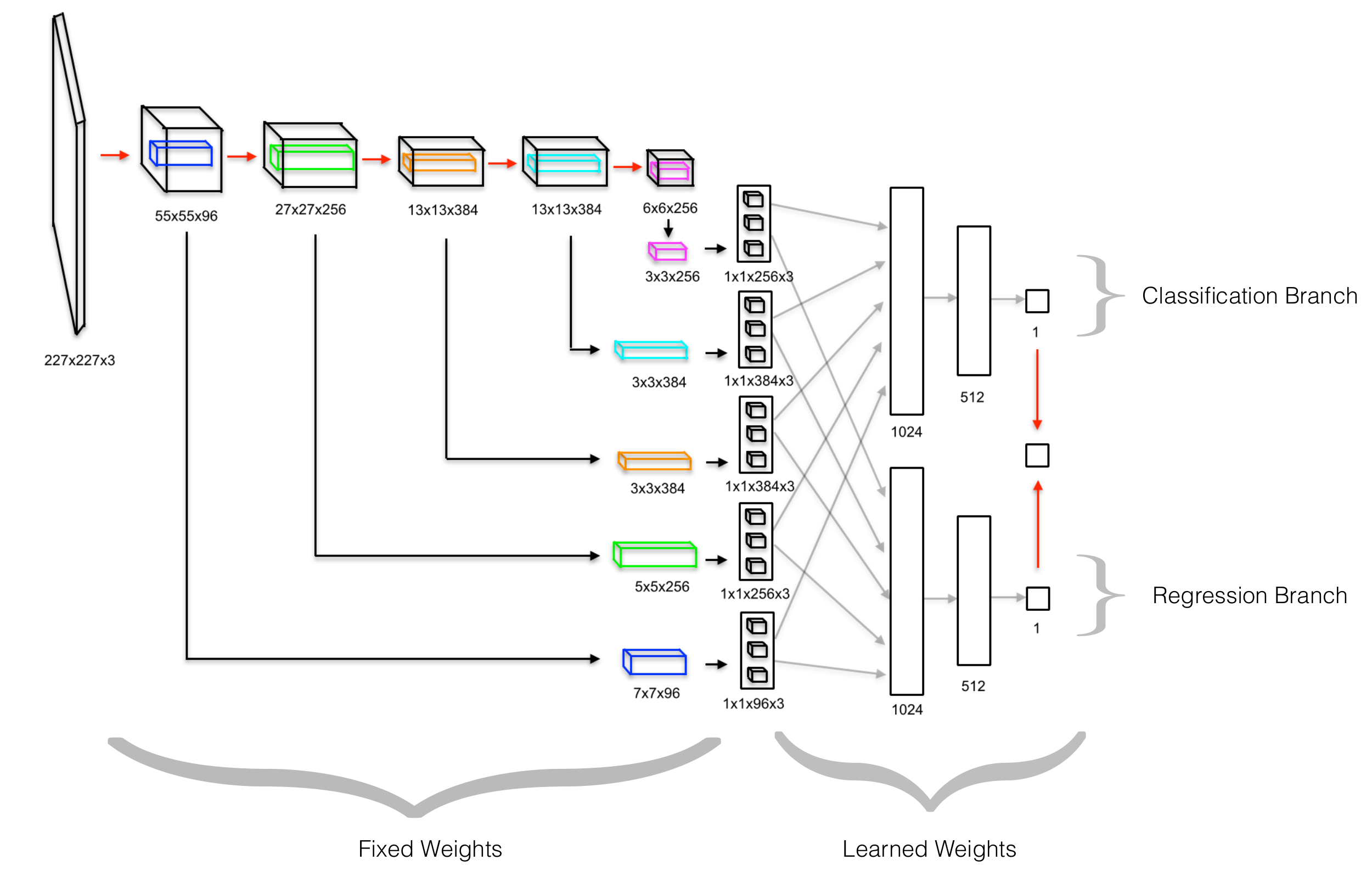}
\end{center}
\setcounter{figure}{4}
   \caption{Detailed illustration of our proposed architecture in a single-scale setting. First, an input patch, centered around the candidate point, goes through five convolutional layers of the {\em KNet}. To extract high-level features, at each convolutional layer we extract a small sub-volume of the feature map around the center point, and perform {\em max}, {\em average}, and {\em center} pooling on this sub-volume. The pooled values feed a bifurcated sub-network. At testing time, the scalar outputs computed from the branches of a bifurcated sub-networks are averaged to produce a final contour prediction.\vspace{-0.2cm}}
\label{fig:ss_arch}
\end{figure*}

\subsection{Multi-Scale Architecture}
\label{sec:multiple}



\captionsetup{labelformat=default}

In the previous section we presented a convolutional architecture for contour prediction utilizing a single scale. However, in practice, we found that a multi-scale approach works much better. In this section, we show how to modify the single-scale architecture so that it can exploit multiple scales simultaneously.

Rather than extracting a patch at a single scale as we did in the previous section, in a multi-scale setting we extract patches around the candidate point for different patch sizes so that they cover different spatial extents of the image. We then resize the patches to fit the KNet input and pump them in parallel through the five convolutional layers. Our high-level features are then built by performing max, average and center pooling in a small sub-volume of the feature map at each convolutional layer and at each scale. This effectively increases the dimensionality of the feature vector by a factor equal to the number of scales compared to the single-scale setting. These pooled features are then connected as before to the two separately-trained network branches. A visualization of our multi-scale architecture is shown in Fig.~\ref{fig:ms_arch}.


\subsection{Implementation Details}


In this section, we describe additional implementation details of our model. Our deep network is implemented using the software library Caffe~\cite{jia2014caffe}.

We use four different scales for our patches. The sizes of these patches are $64 \times 64, 128 \times 128, 196 \times 196$ and a full-sized image. All of the patches are then resized to the {\em KNet} input dimensions of $227 \times 227$.
  
When extracting high-level features from the convolutional layers of {\em KNet}, we use sub-volumes of convolutional feature maps having spatial sizes $7\times 7, 5 \times 5, 3 \times 3, 3 \times 3$, and $3 \times 3$ for the convolutional layers $1,2,3,4,5$, respectively. Note that we shrink the size of the subvolume as we go deeper in the network since the feature maps get smaller due to pooling. Our choice of subvolume sizes is made to ensure we are roughly considering the same spatial extent of the original image at each layer.

As illustrated in Fig.~\ref{fig:ss_arch}, during the training the weights in the convolutional layers are fixed and only the weights in the fully connected layers of the two branches are learned. 

To train our model we use the learning rate of $0.1$,  the dropout fraction of $0.5$,  $50$ number of epochs, and the size of the batch equal to $100$.

As described earlier, to train classification and regression branches we sample $80,000$ examples with binary labels. We also generate continuous labels for these $80,000$ examples. In addition, we sample a hold-out dataset of size $40,000$. This hold-out dataset is used for the hard-positive mining step~\cite{malisiewicz-iccv11}.

For the first $25$ epochs we train classification and regression branches independently on the original $80,000$ sample training dataset. After the first $25$ epochs, we test both branches on the hold-out dataset and detect false negative predictions made by each branch. We then use these false negative examples along with the same number of randomly selected true negative examples to augment our original $80,000$ training dataset. For the remaining $25$ epochs, we train both branches on this augmented dataset. 

The motivation for the hard-positive mining step is to reduce the number of false negative predictions produced by both branches. By augmenting the original $80,000$ sized training data with false negative examples, we are forcing both branches to focus on {\em hard positive} examples, and thus, effectively reducing the number of false negative predictions.


\section{Experiments}

\captionsetup{labelformat=empty}

\begin{figure*}
\centering
\begin{minipage}[b]{.19\textwidth}
\includegraphics[width=1\linewidth]{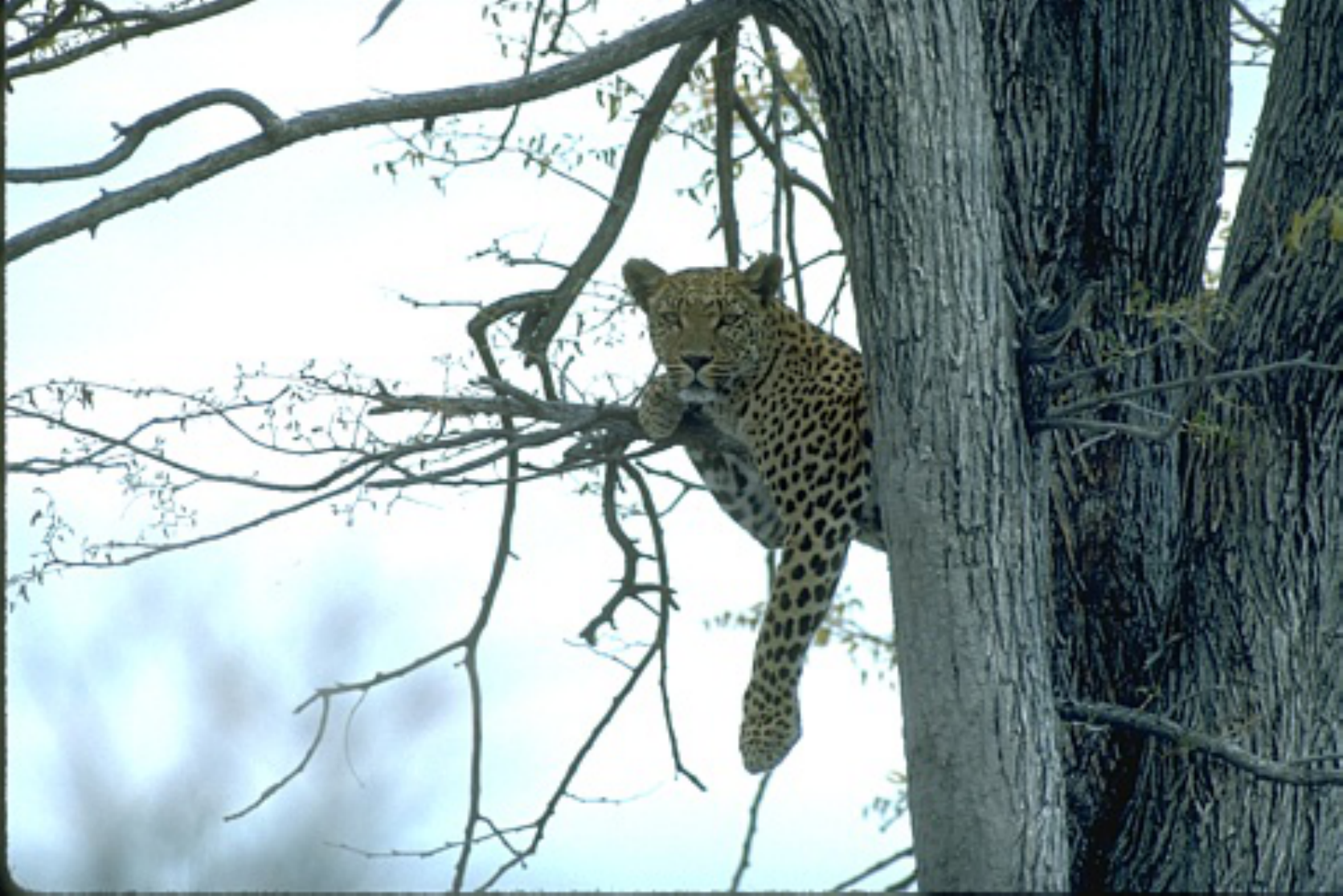}
\end{minipage}
\begin{minipage}[b]{.19\textwidth}
\includegraphics[width=1\linewidth]{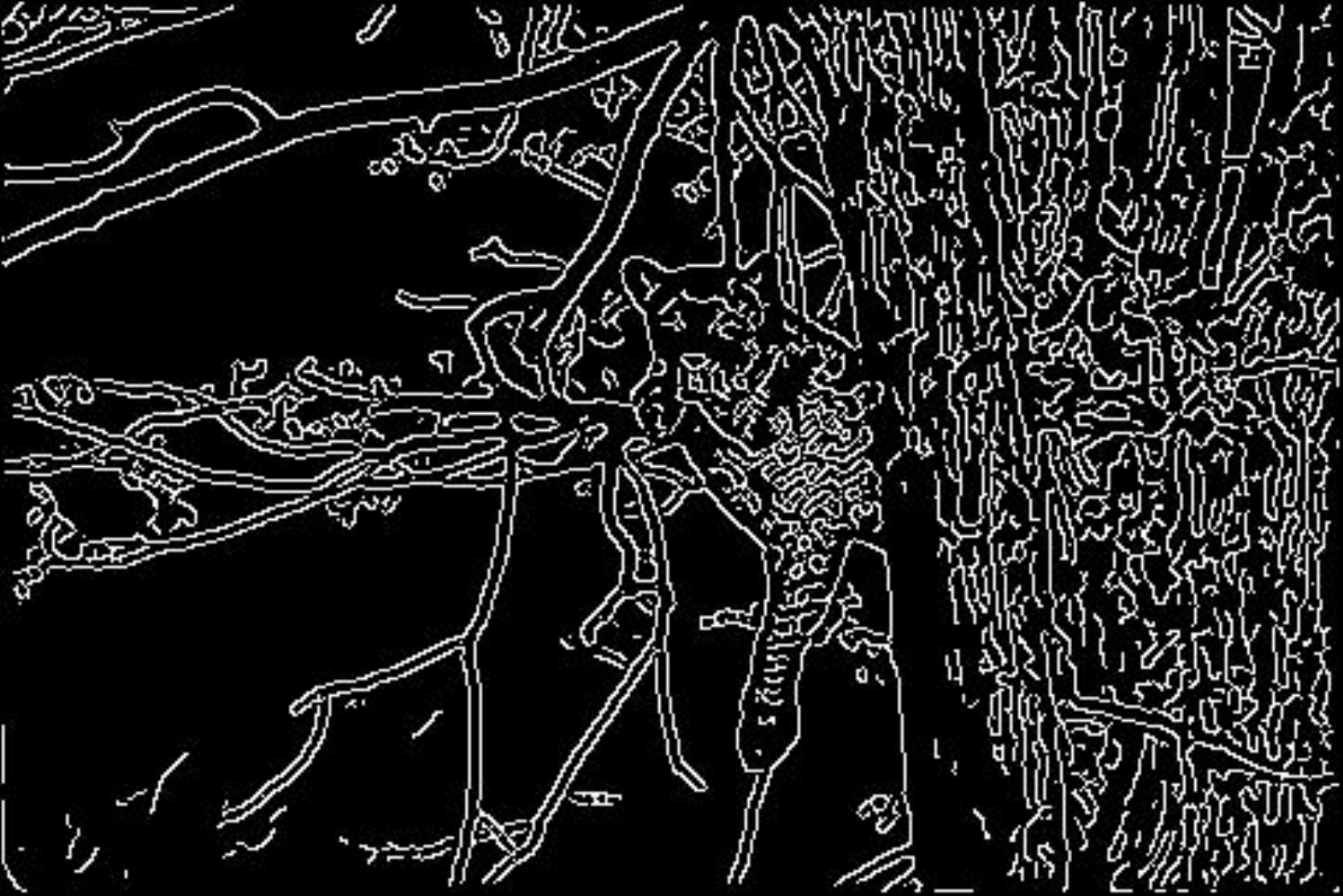}
\end{minipage}
\begin{minipage}[b]{.19\textwidth}
\includegraphics[width=1\linewidth]{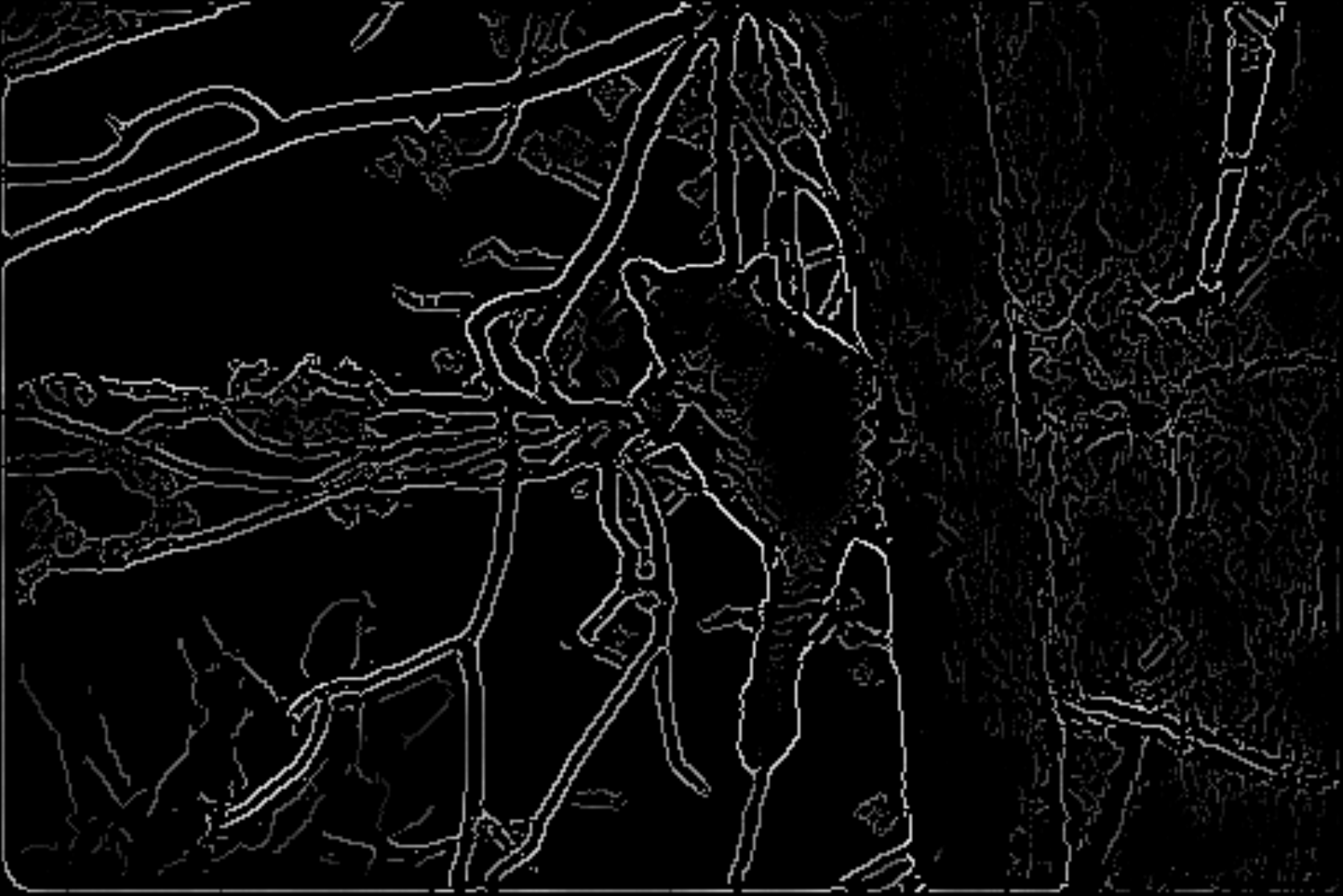}
\end{minipage}
\begin{minipage}[b]{.19\textwidth}
\includegraphics[width=1\linewidth]{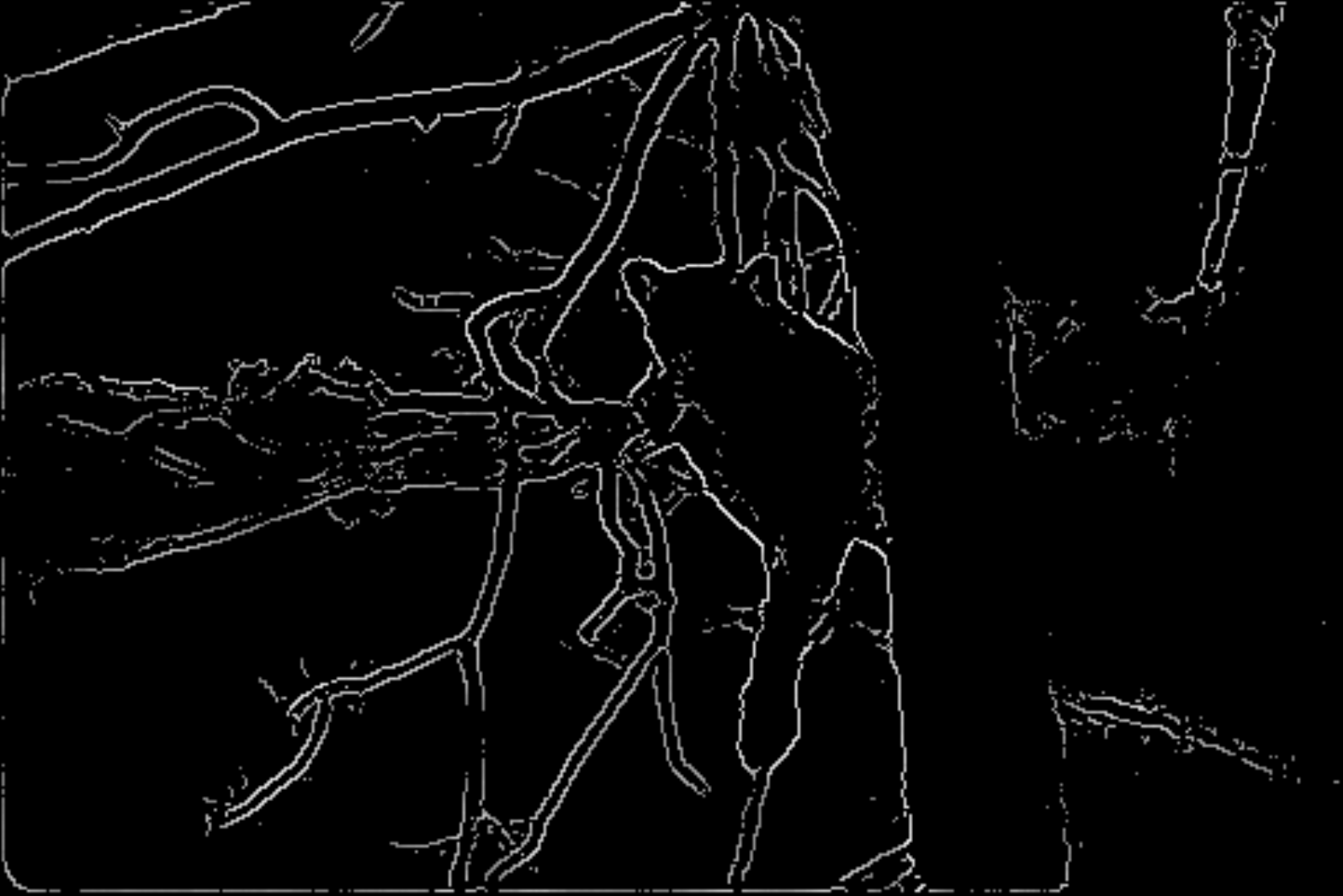}
\end{minipage}
\begin{minipage}[b]{.19\textwidth}
\includegraphics[width=1\linewidth]{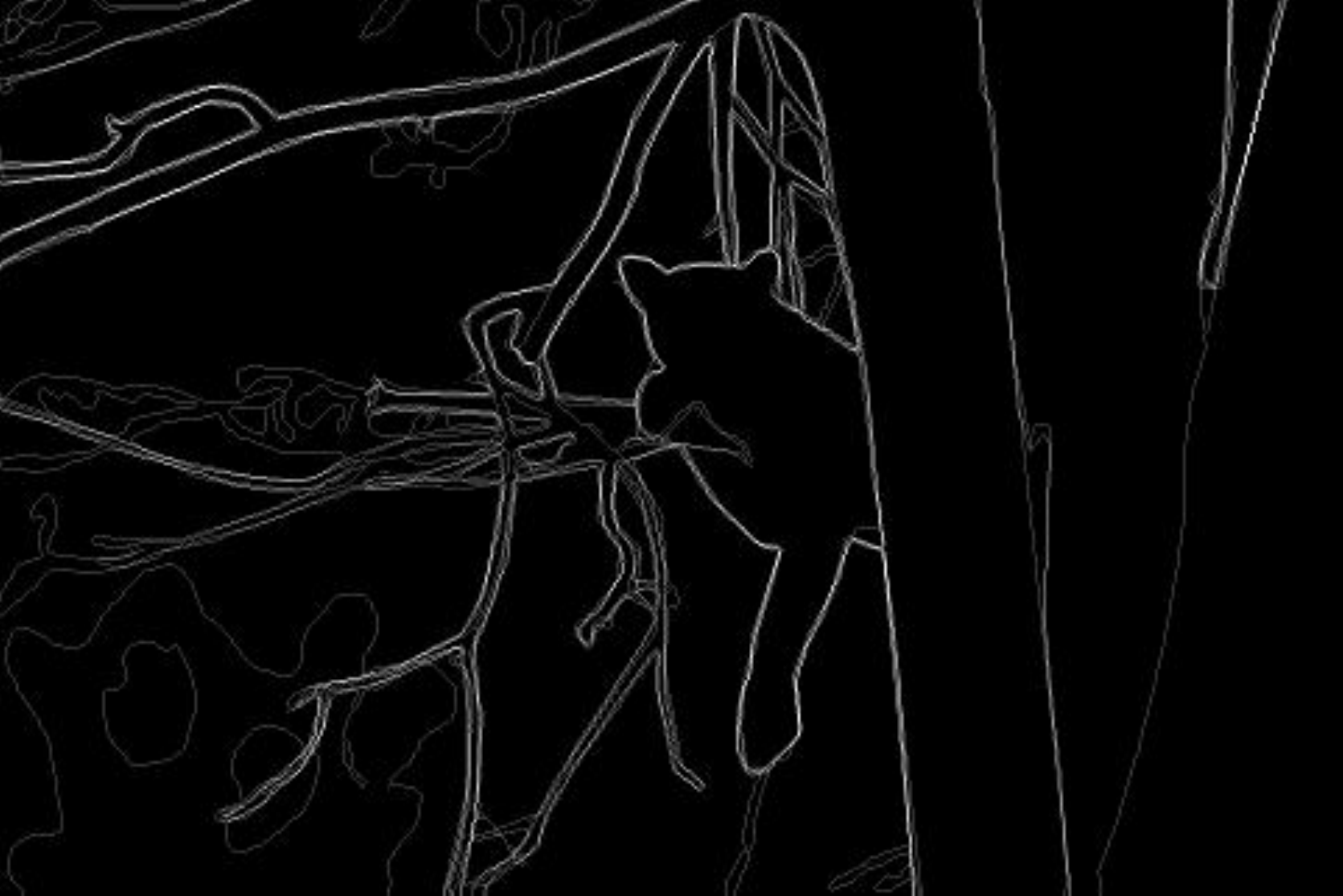}
\end{minipage}

\begin{minipage}[b]{.19\textwidth}
\includegraphics[width=1\linewidth]{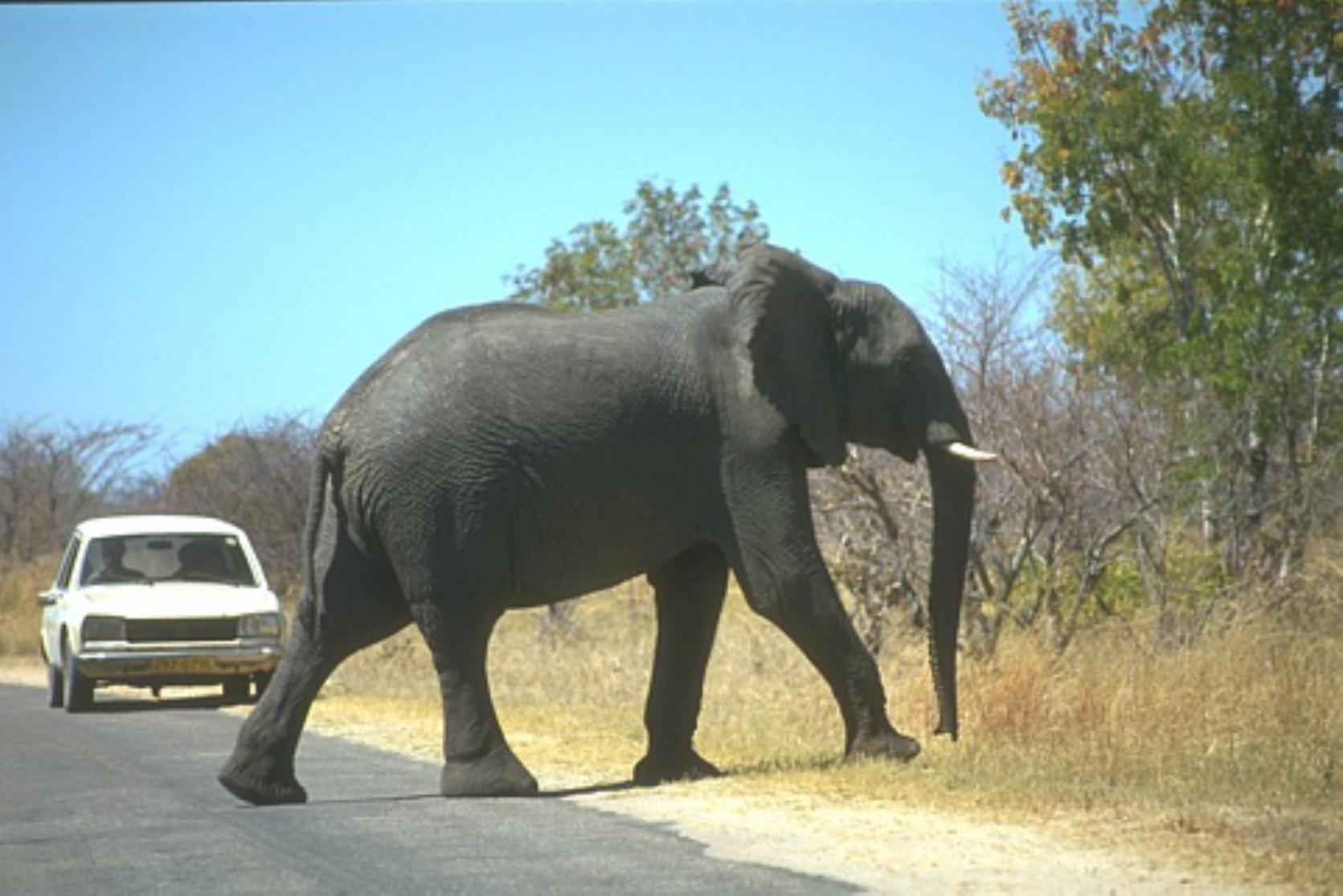}
\end{minipage}
\begin{minipage}[b]{.19\textwidth}
\includegraphics[width=1\linewidth]{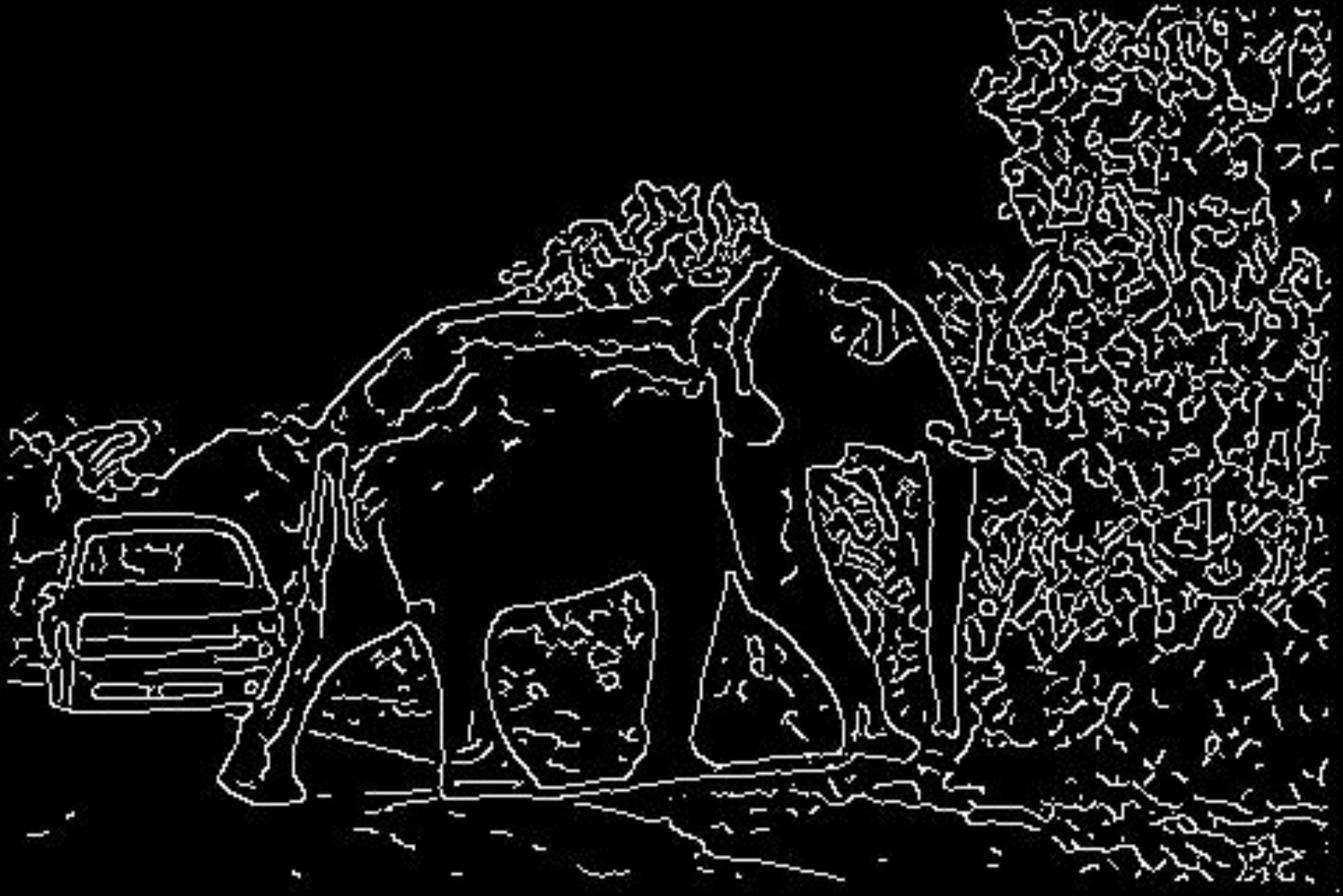}
\end{minipage}
\begin{minipage}[b]{.19\textwidth}
\includegraphics[width=1\linewidth]{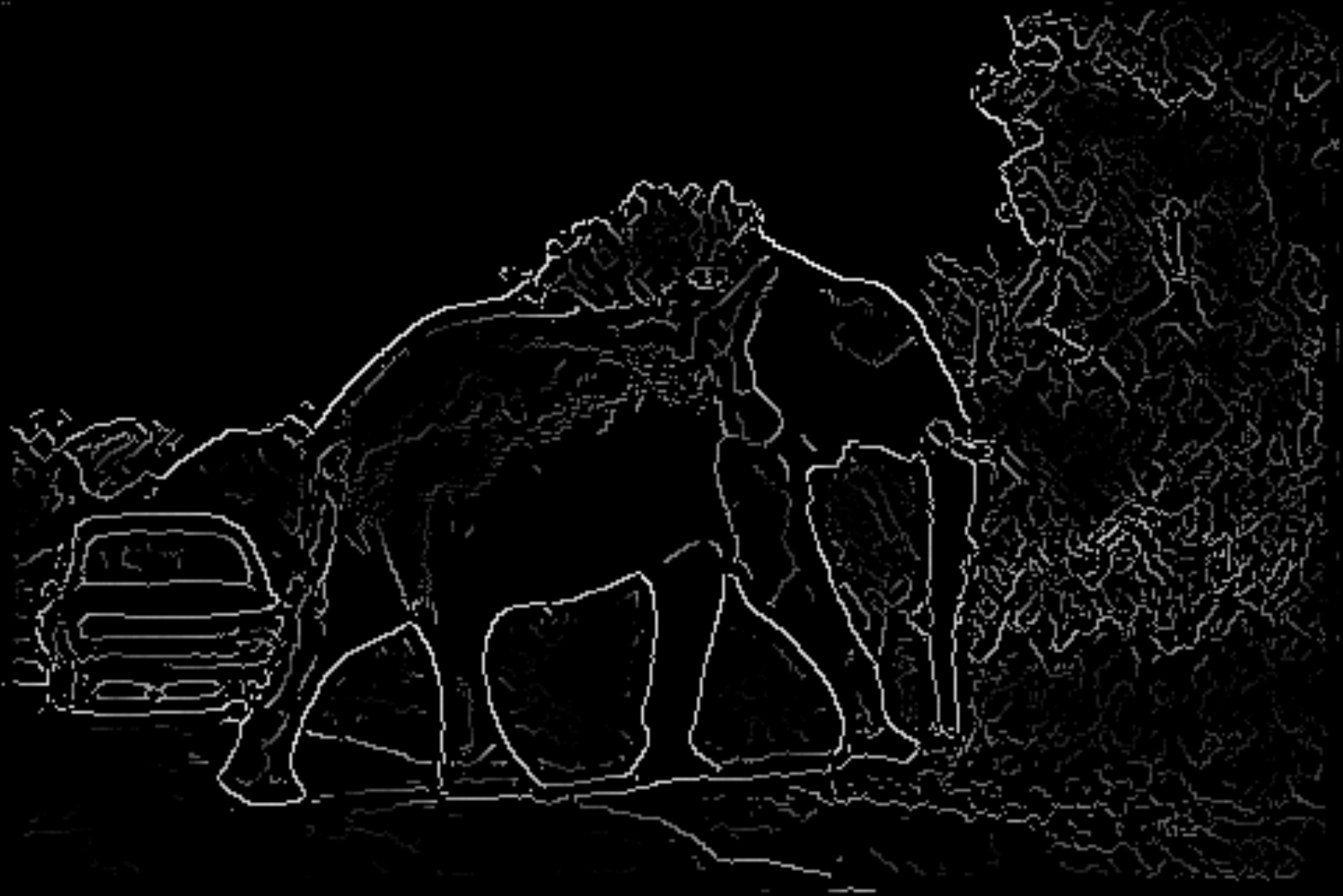}
\end{minipage}
\begin{minipage}[b]{.19\textwidth}
\includegraphics[width=1\linewidth]{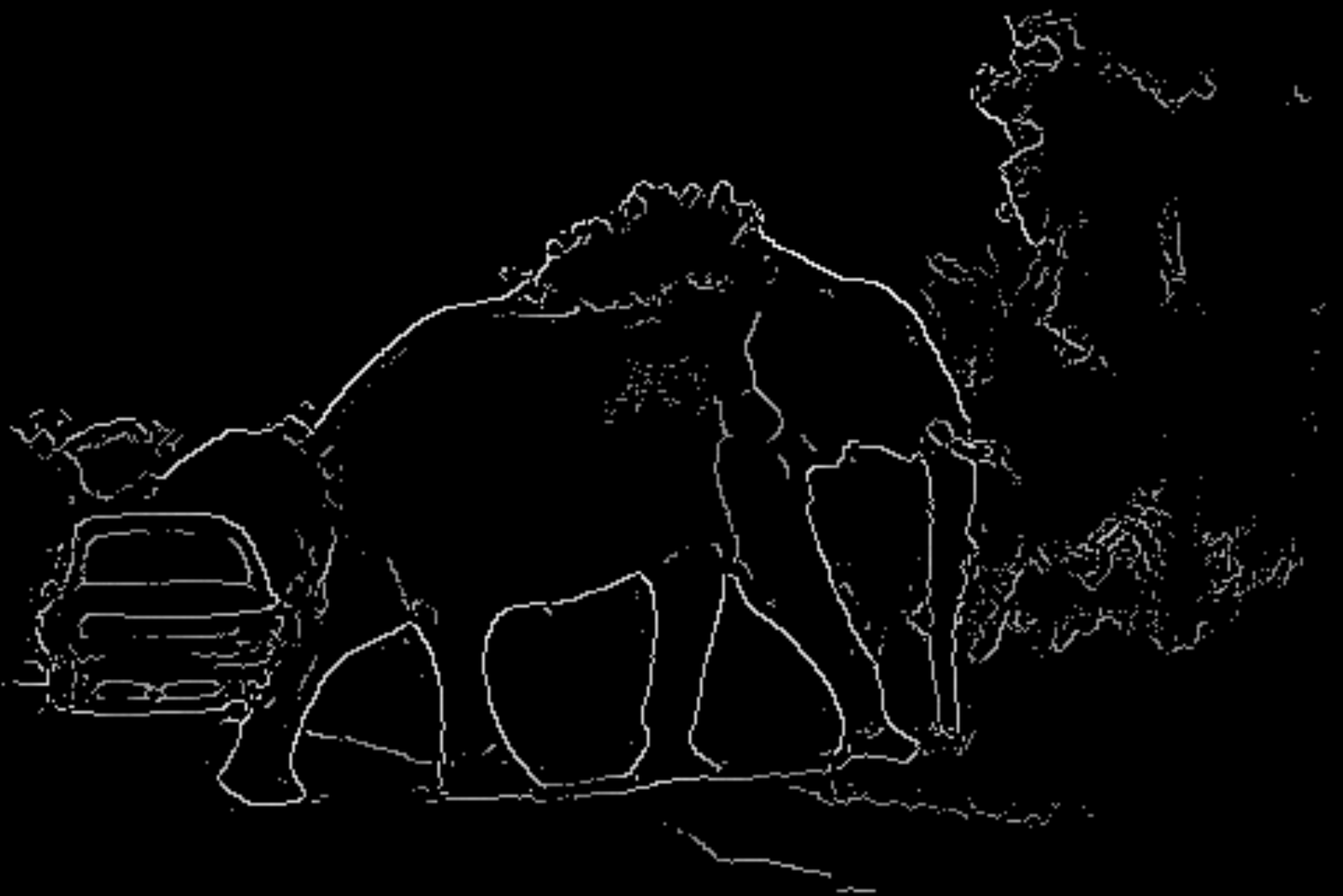}
\end{minipage}
\begin{minipage}[b]{.19\textwidth}
\includegraphics[width=1\linewidth]{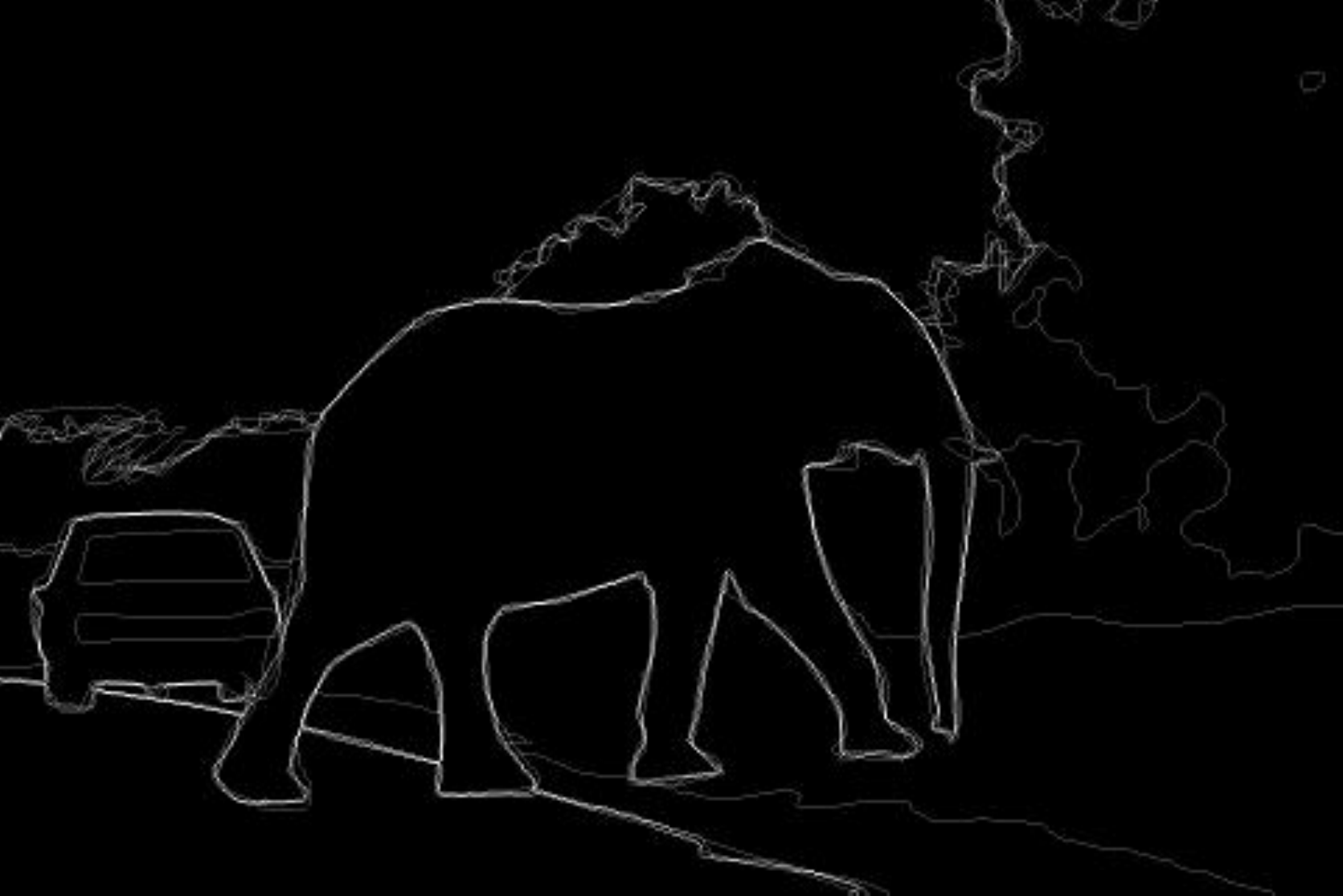}
\end{minipage}

\begin{minipage}[b]{.19\textwidth}
\includegraphics[width=1\linewidth]{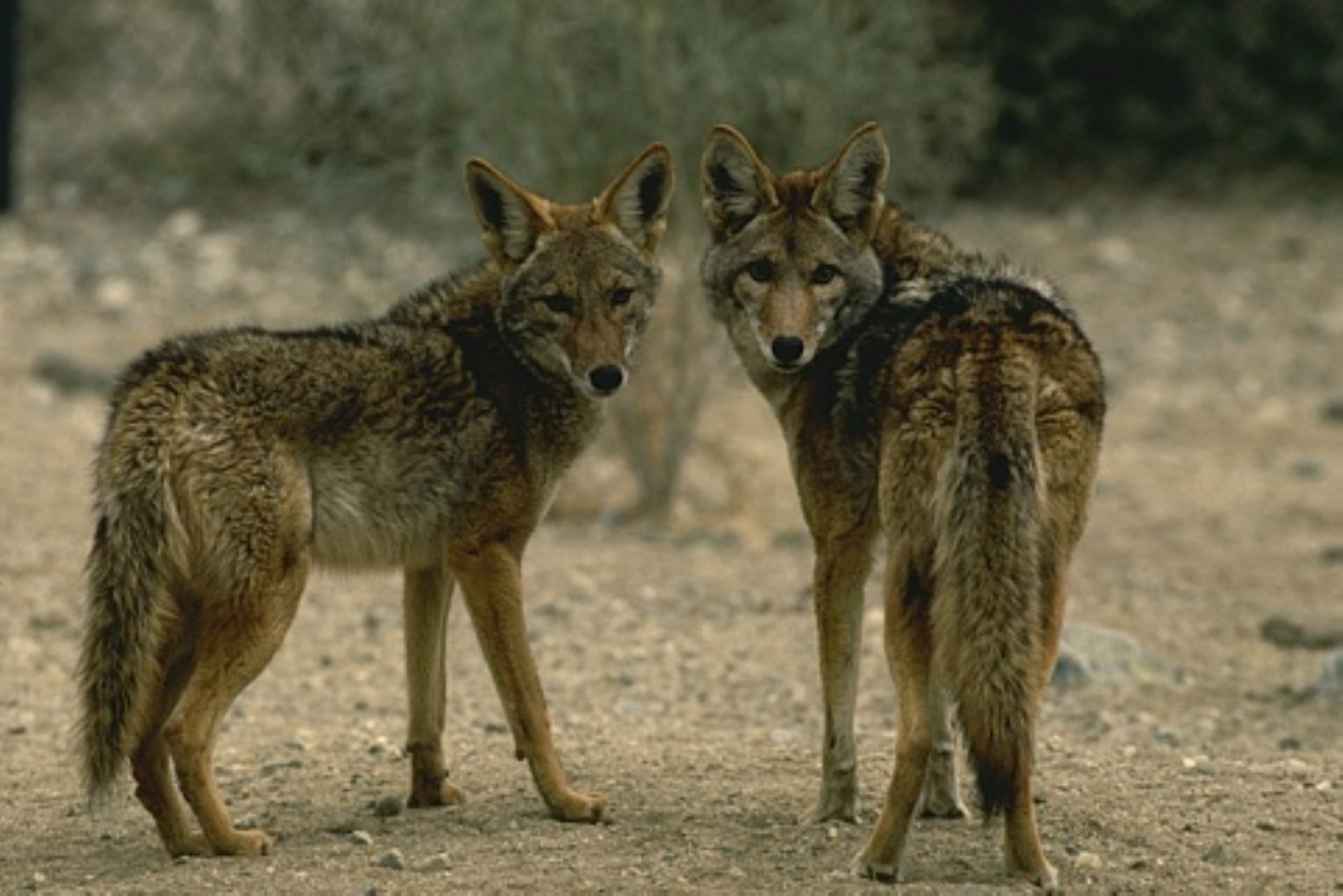}
\end{minipage}
\begin{minipage}[b]{.19\textwidth}
\includegraphics[width=1\linewidth]{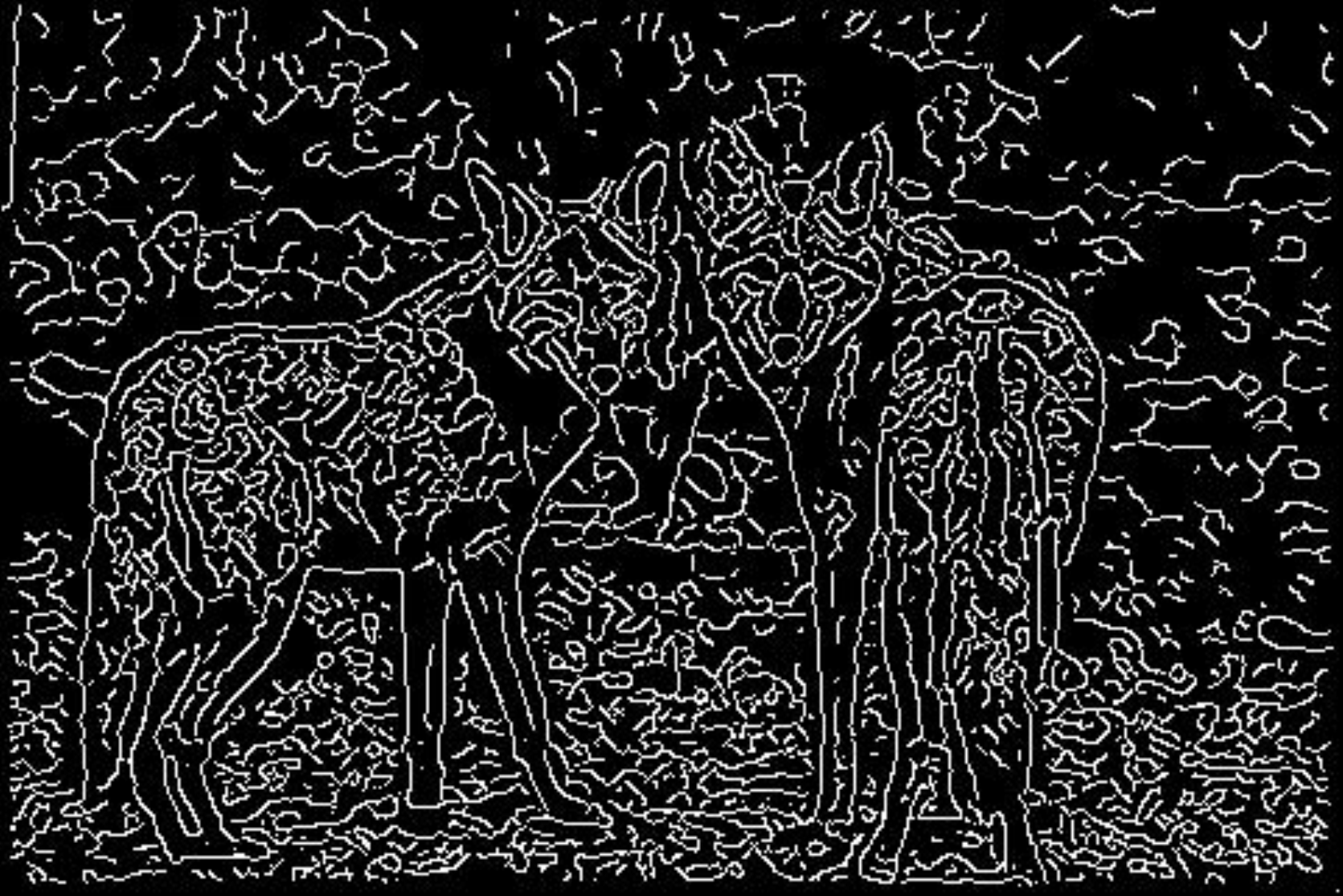}
\end{minipage}
\begin{minipage}[b]{.19\textwidth}
\includegraphics[width=1\linewidth]{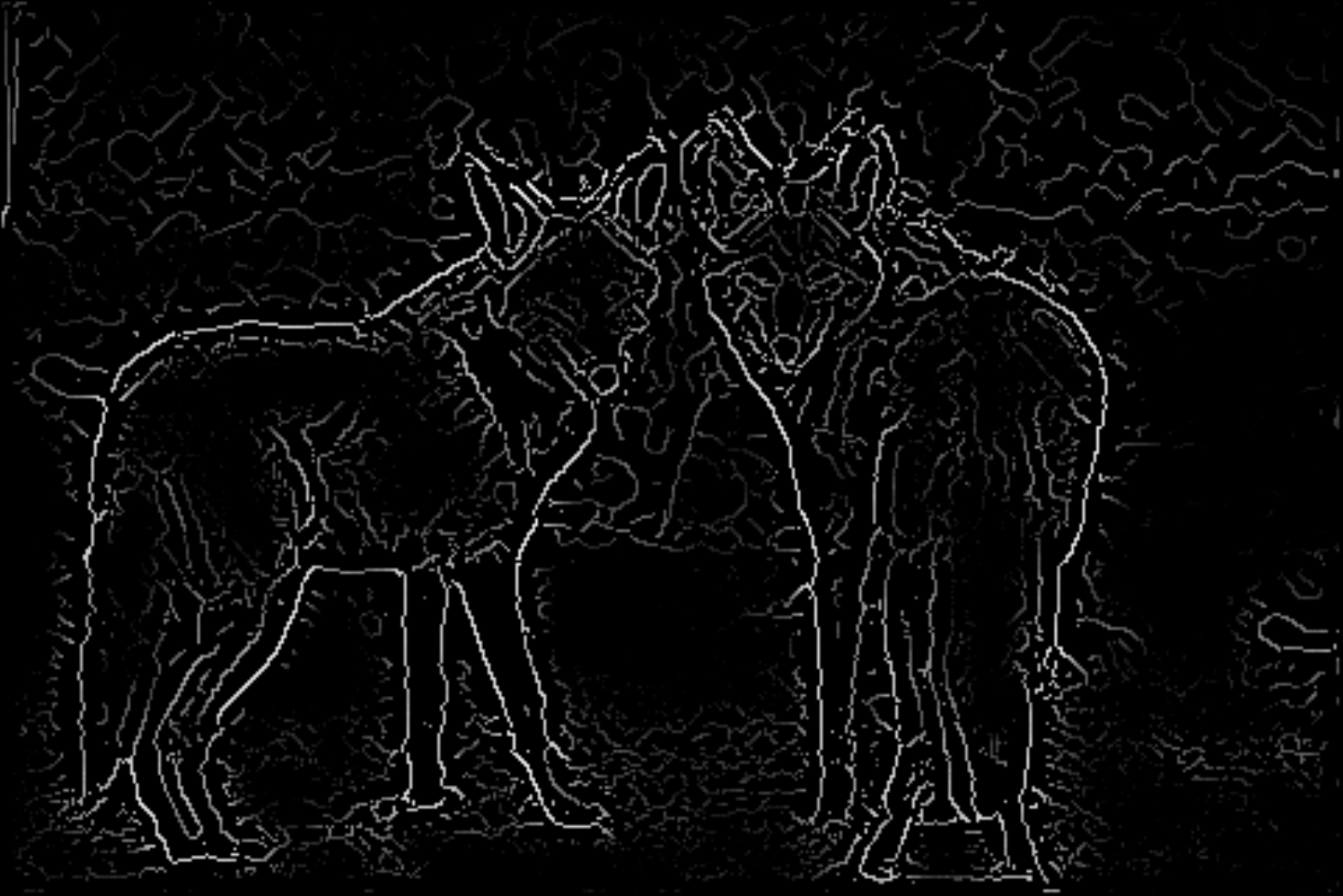}
\end{minipage}
\begin{minipage}[b]{.19\textwidth}
\includegraphics[width=1\linewidth]{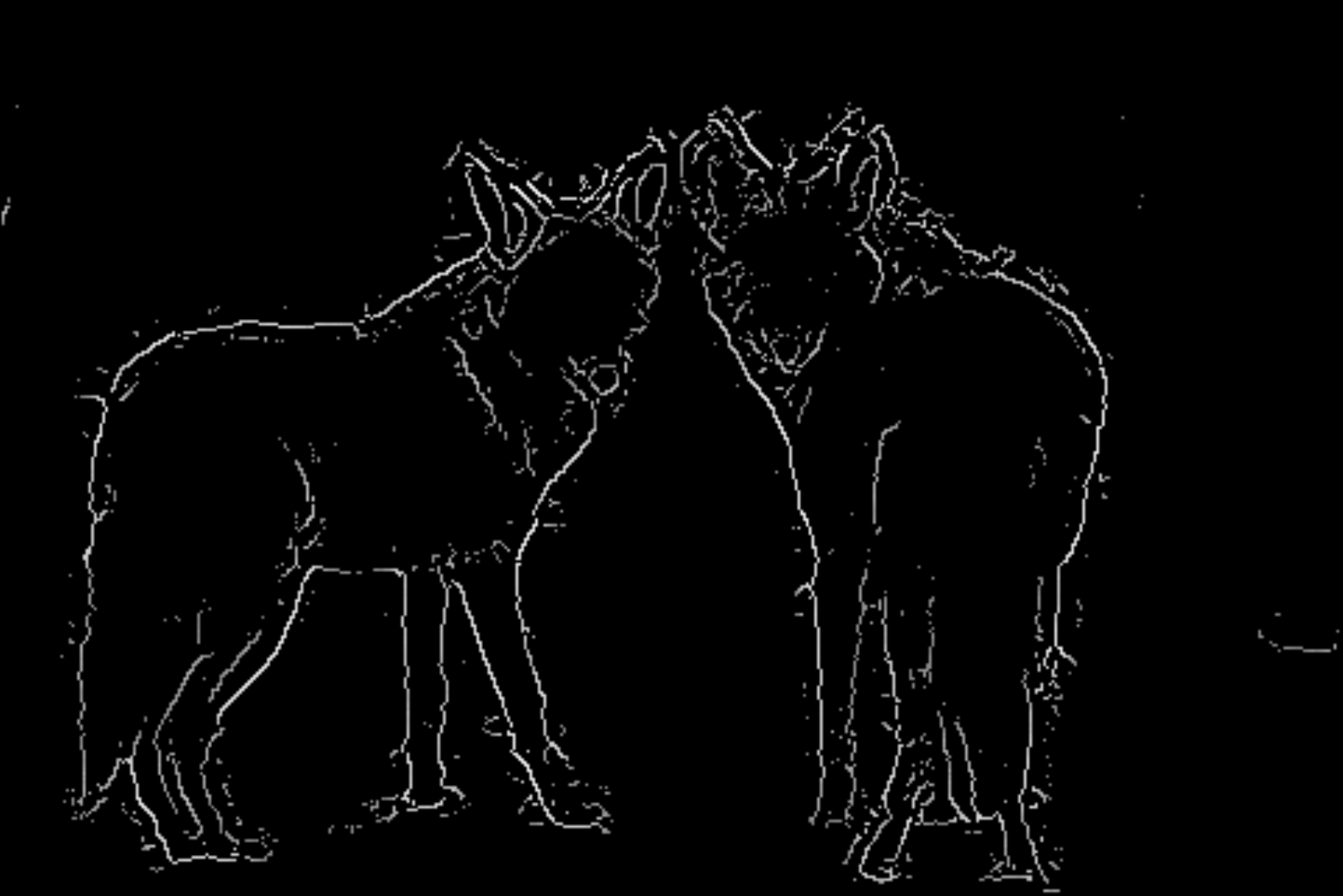}
\end{minipage}
\begin{minipage}[b]{.19\textwidth}
\includegraphics[width=1\linewidth]{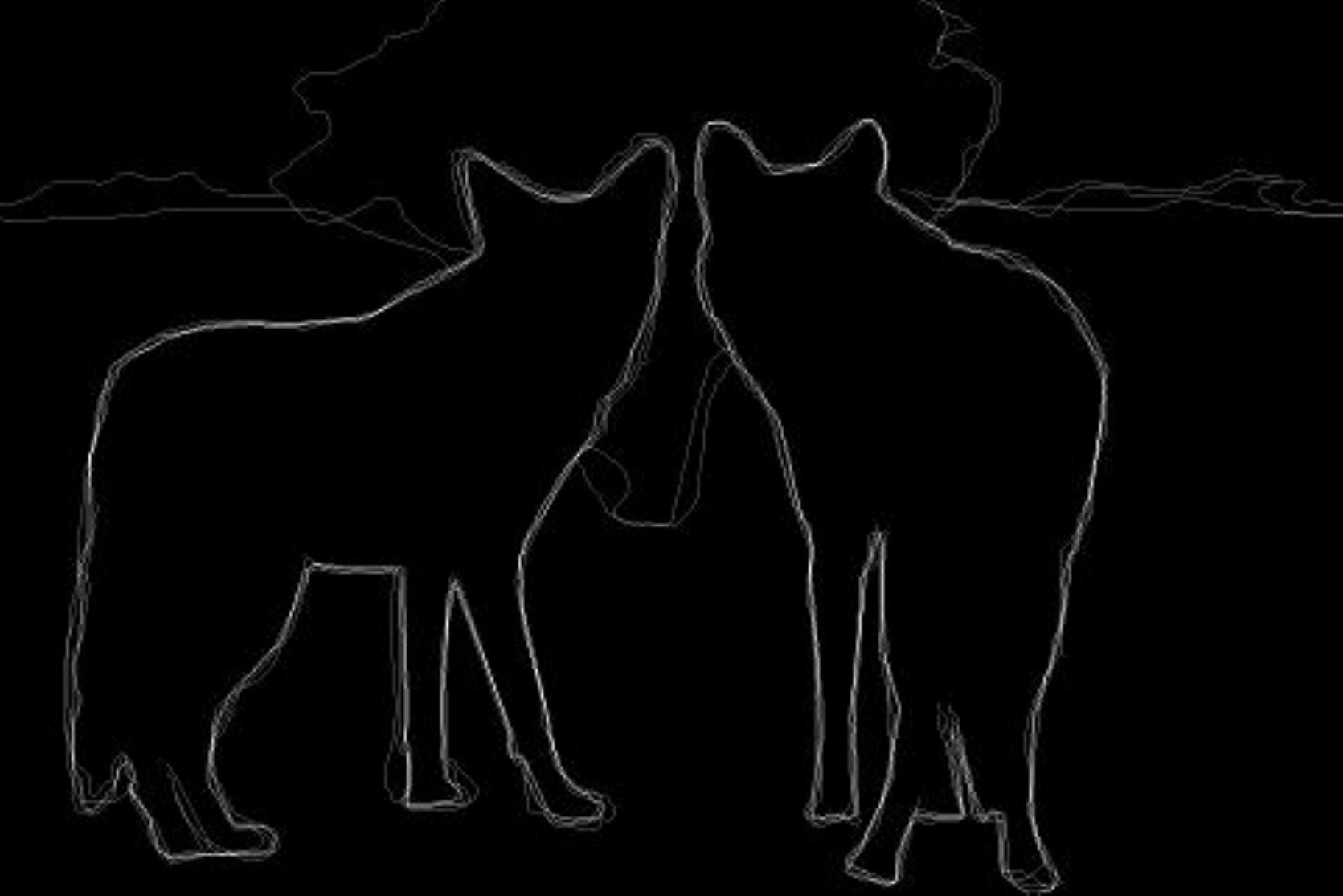}
\end{minipage}


\begin{minipage}[b]{.19\textwidth}
\includegraphics[width=1\linewidth]{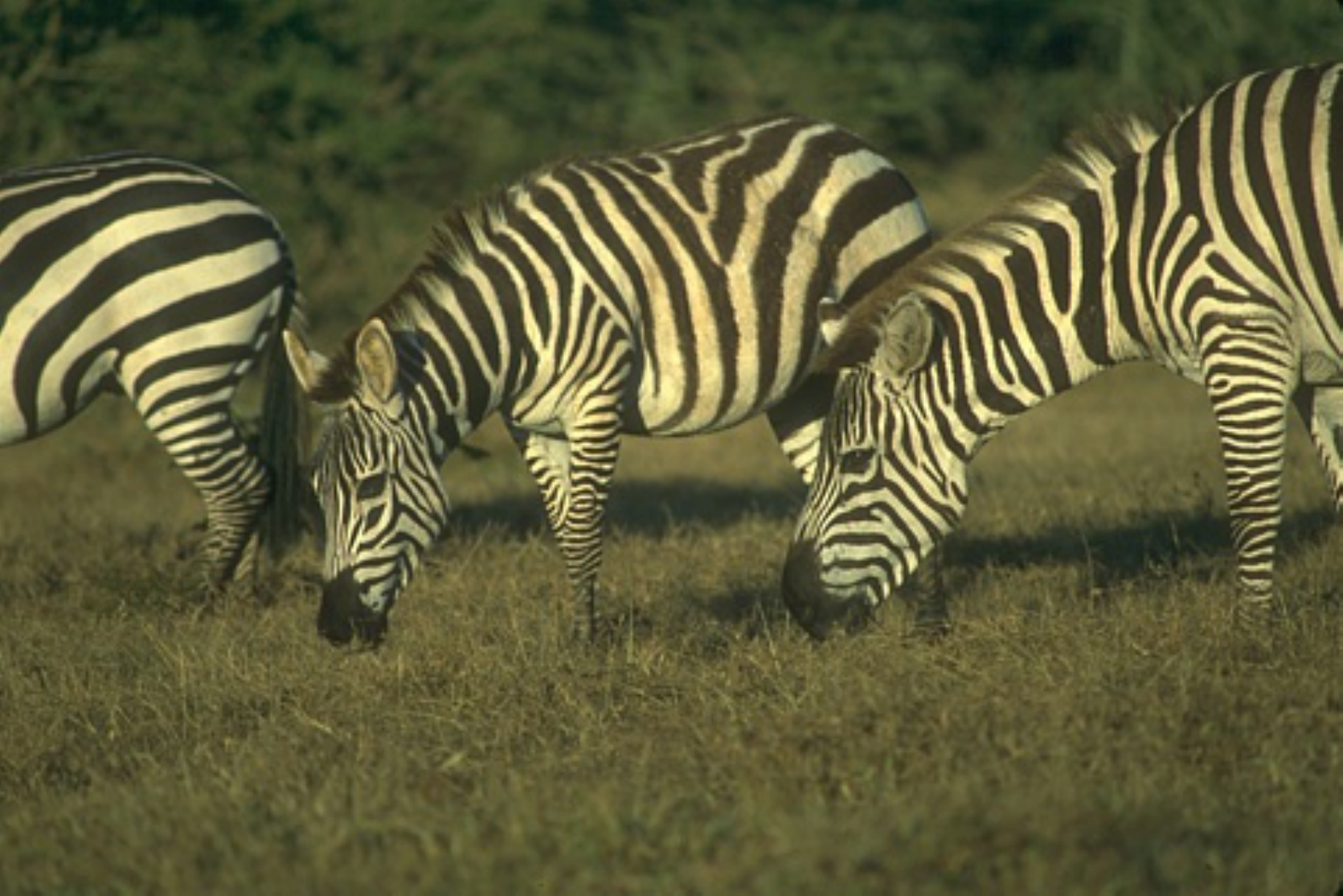}
\caption{{\small Input Image}}
\end{minipage}
\begin{minipage}[b]{.19\textwidth}
\includegraphics[width=1\linewidth]{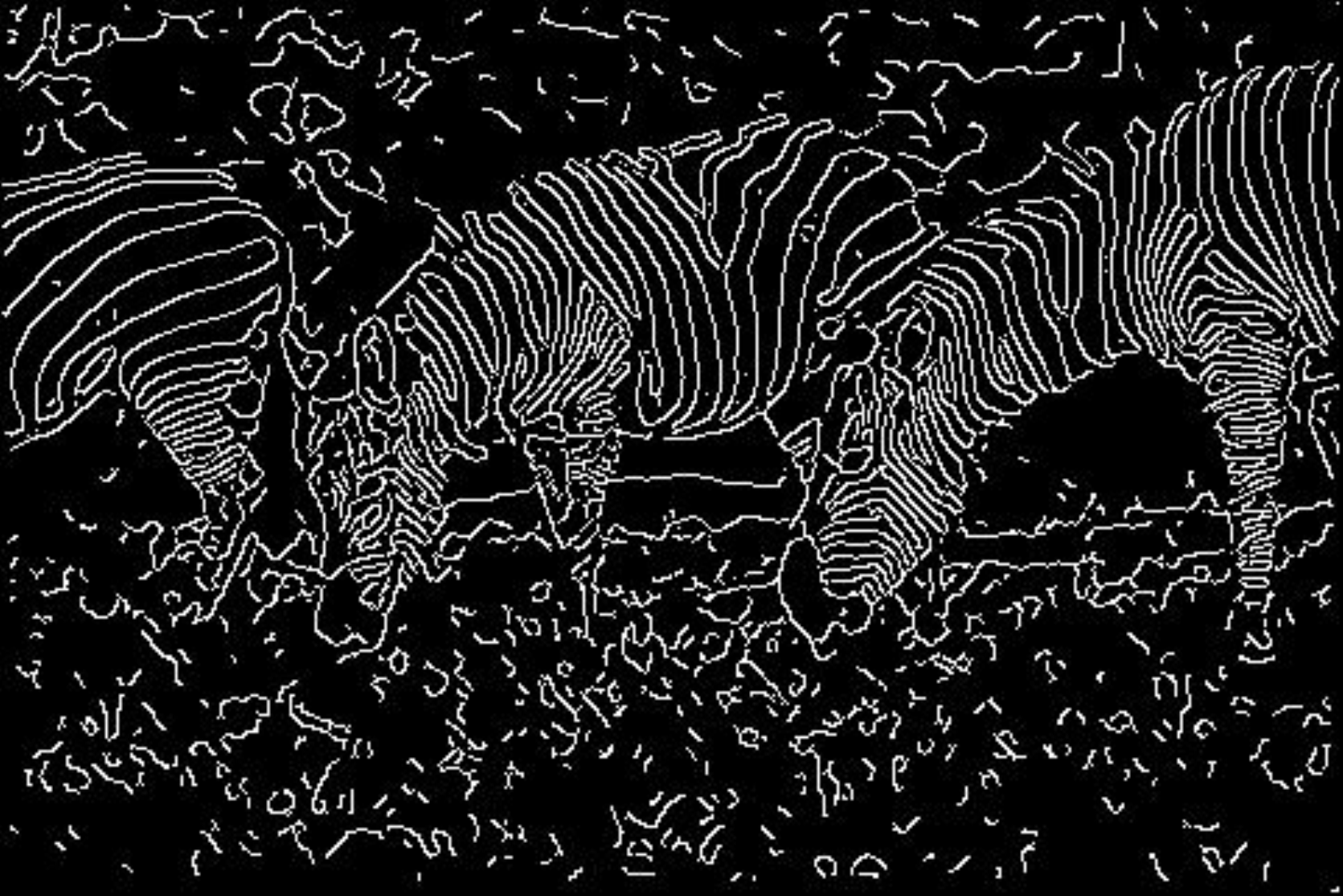}
\caption{{\small Canny Edges}}
\end{minipage}
\begin{minipage}[b]{.19\textwidth}
\includegraphics[width=1\linewidth]{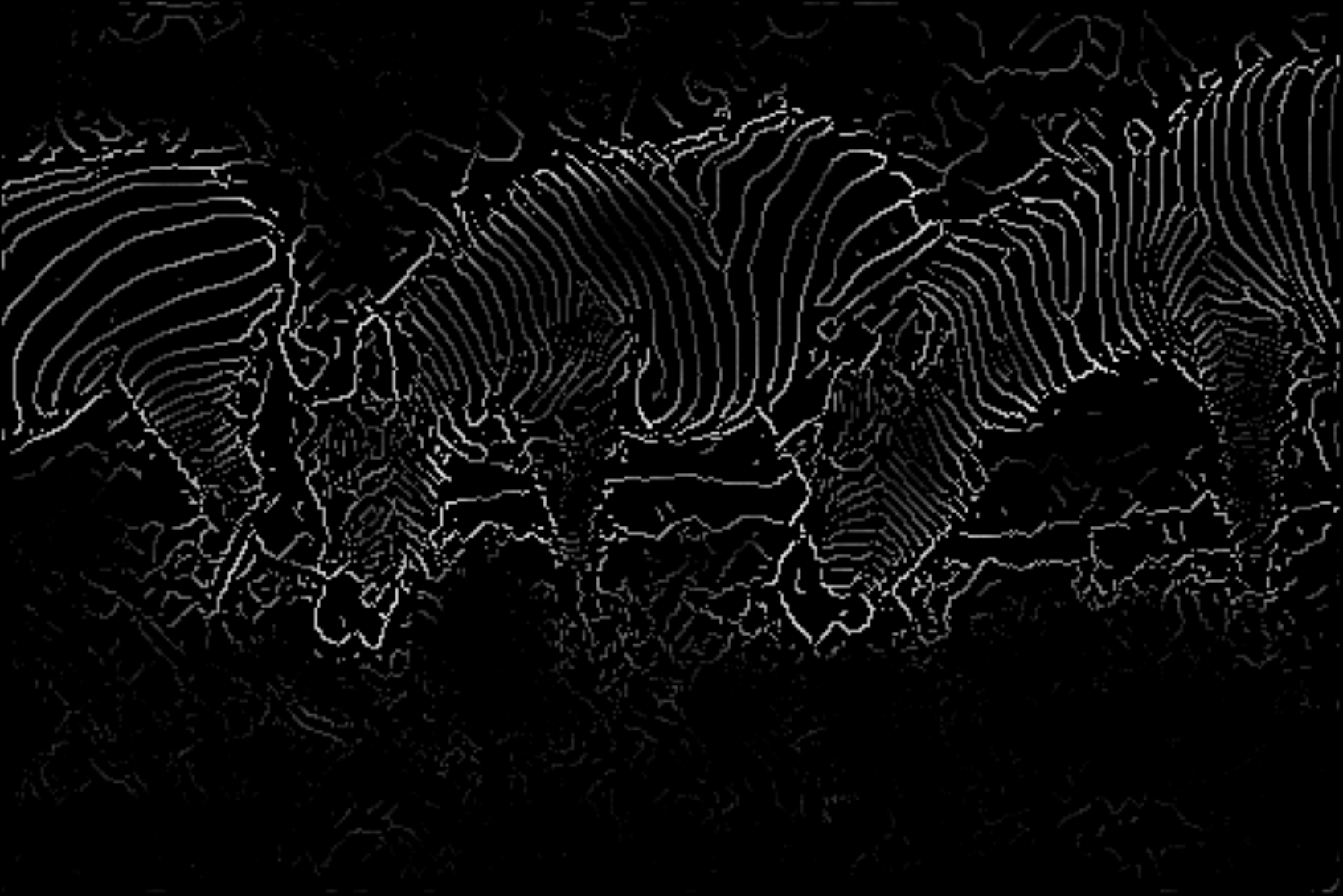}
\caption{{\small Raw DeepEdges}}
\end{minipage}
\begin{minipage}[b]{.19\textwidth}
\includegraphics[width=1\linewidth]{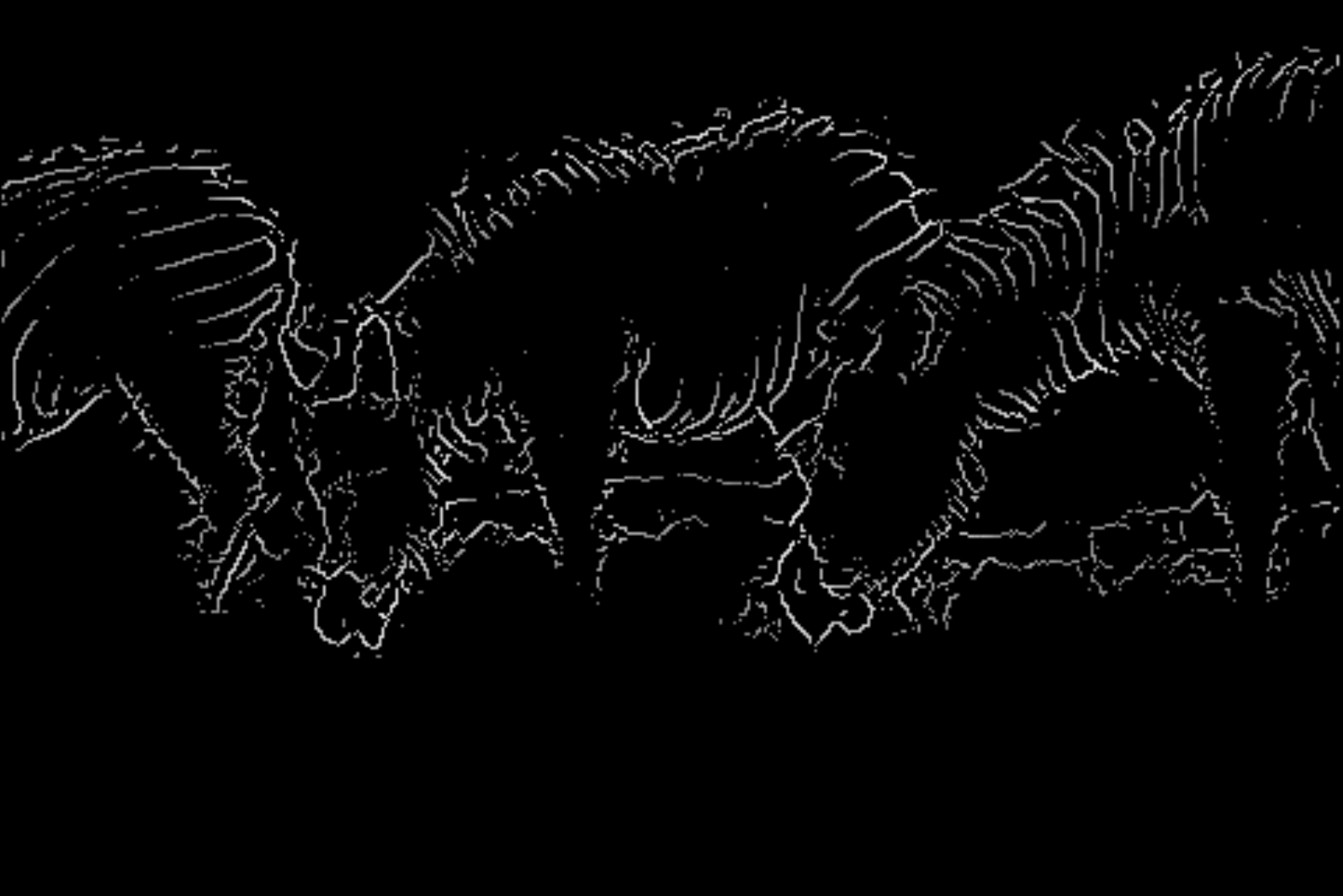}
\caption{{\small Thresholded DeepEdges}}
\end{minipage}
\begin{minipage}[b]{.19\textwidth}
\includegraphics[width=1\linewidth]{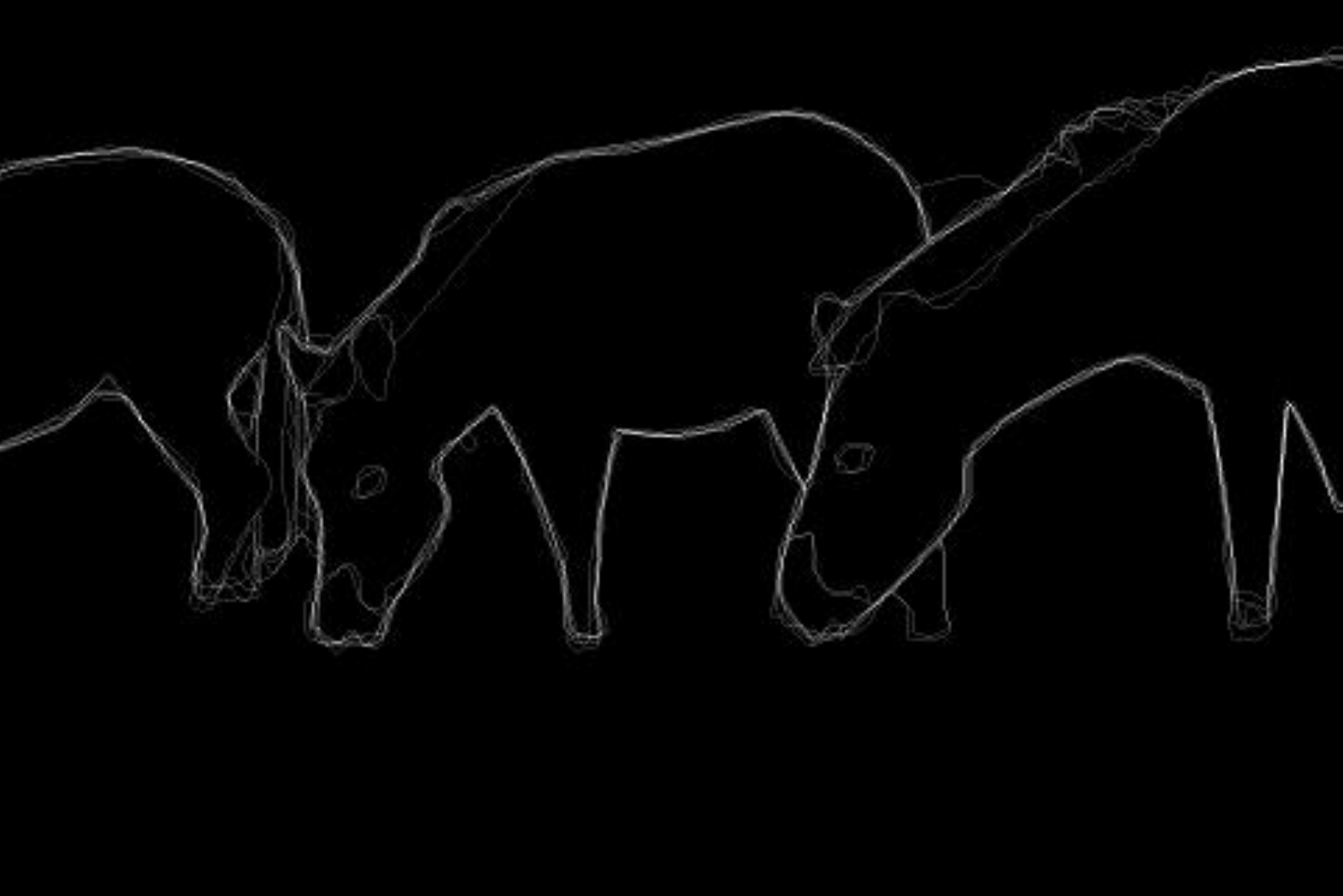}
\caption{{\small Ground Truth Edges}}
\end{minipage}
\captionsetup{labelformat=default}
\setcounter{figure}{5}
    \caption{Qualitative results produced by our method. Notice how our method learns to distinguish between strong and weak contours. For instance, in the last row of predictions, contours corresponding to zebra stripes are assigned much lower probabilities than contours that correspond to the actual object boundaries separating the zebras from the background.}
    \label{qualitative_results}
\end{figure*}

\captionsetup{labelformat=default}


In this section, we present our results on the BSDS500 dataset~\cite{MartinFTM01}, which is arguably the most established benchmark for contour detection. This dataset contains $200$ training images, $100$ validation images, and $200$ test images. Contour detection accuracy is evaluated using three standard measures: fixed contour threshold (ODS), per-image best threshold (OIS), and average precision (AP).

In section~\ref{sec:comparison} we quantitatively compare our approach to the state-of-the-art. In sections~\ref{sec:expmultiplescale}-\ref{sec:expobj} we study how the performance of our system changes as we modify some of the architecture choices (number of scales, feature maps, pooling scheme, training objective). This will cast additional insight into the factors that critically contribute to the high accuracy of our system.


\subsection{Comparison with Other Methods}
\label{sec:comparison}

We compare the results produced by our approach and previously proposed contour detection methods. 
Table~\ref{results_methods} summarizes the results. We note that our algorithm achieves contour detection accuracy that is higher or equal to state-of-the-art results according to two of the three metrics. 

Fig.~\ref{fig:isoF} shows the precision and recall curve for the methods considered in our comparison. It also lists the F-score for each method (in the legend). We observe that there is the accuracy margin separating our approach from prior techniques. In particular, for low-recall our method achieves almost perfect precision rate. It also produces state-of-the-art F-score. 

%

  \begin{table}
    \begin{center}
    \begin{tabular}{ | c | c | c | c |}
    \hline
    Method & ODS & OIS & AP \\ \hline\hline
    	Felz., Hutt.~\cite{Felzenszwalb:2004:EGI:981793.981796} & 0.610 & 0.640 & 0.560 \\ \hline
	Mean Shift~\cite{Comaniciu02meanshift:} & 0.640 & 0.680 & 0.560 \\ \hline
	Ncuts~\cite{Shi97normalizedcuts} & 0.640 & 0.680 & 0.450\\ \hline
	SCT~\cite{MYP:ACCV:2014} & 0.710 & 0.720 & 0.740 \\ \hline
	gPb-owt-ucm~\cite{Arbelaez:2011:CDH:1963053.1963088} & 0.726 & 0.757 & 0.696\\ \hline
	Sketch Tokens~\cite{LimCVPR13SketchTokens} & 0.727 & 0.746 & 0.780\\ \hline
	PMI~\cite{crisp_boundaries} & 0.737 & 0.771 & 0.783\\ \hline
	DeepNet~\cite{kivinen2014visual} & 0.738 & 0.759 & 0.758 \\ \hline
	SCG~\cite{ren_nips12} & 0.739 & 0.758 & 0.773 \\ \hline
	SE~\cite{Dollar2015PAMI} & 0.746 & 0.767 & 0.803\\ \hline
	MCG~\cite{cArbelaez14} & 0.747 & \bf 0.779 & 0.759\\ \hline
	$N^4$-fields~\cite{DBLP:journals/corr/GaninL14} & \bf 0.753 & 0.769 & 0.784\\ \hline
	\bf DeepEdge & \bf 0.753 & 0.772 & \bf 0.807\\
	
    \hline
    \end{tabular}
    \end{center}
    \caption{Edge detection results on the BSDS500 benchmark. Our DeepEdge method achieves state-of-the-art contour detections results according to both F-score and AP metrics.}
    \label{results_methods}
   \end{table}


\subsection{Single Scale versus Multiple Scales}
\label{sec:expmultiplescale}

In this section we study the benefits of a multi-scale architecture. Results in Table~\ref{results_scale} report accuracy for different numbers and choices of scales. The first four rows in the table illustrate the results achieved using a single-scale approach. Specifically, these four cases show performance obtained when training and testing our system  with an input patch of size $64 \times 64, 128 \times 128, 196 \times 196 $ or a full-sized image, respectively. Note that adding information from multiple scales leads to significantly higher F-scores and higher average precisions. Thus, these results suggest that a multi-scale approach is highly advantageous in comparison to a single scale setting.


  \begin{table}
  	\begin{center}
    \begin{tabular}{ | c | c | c | c |}
    \hline
    Scale & ODS & OIS & AP\\ \hline\hline
    64 & 0.71 & 0.73 & 0.76 \\ \hline
	128 & 0.72 & 0.74 & 0.78 \\ \hline
	196 & 0.71 & 0.73 & 0.76 \\ \hline
	Full Image & 0.67 & 0.69 & 0.57\\ \hline
	64, 128 & 0.72 & 0.75 & 0.78\\ \hline
	64, 128,196 & 0.72 & 0.75 & 0.78 \\ \hline
	64,128,196,Full Image & \bf 0.75 & \bf 0.77 & \bf 0.81 \\
	
    \hline
    \end{tabular}
    \end{center}
    
    \caption{Results illustrating the effect of using a multi-scale architecture. Considering multiple scales for contour detection yields significantly higher accuracy relative to a single scale approach.\vspace{-0.6cm}}
    \label{results_scale}
   \end{table}

\subsection{Advantages of Higher-Level Features}

In this section, we examine the validity of our earlier claim that higher-level object-features enhance contour detection accuracy. In Table~\ref{results_ind_conv}, we present individual contour detection results using features from the different convolutional layers of {\em KNet}. Note that the $4^{th}$ convolutional layer produces the most effective features when considering one layer at a time. From our earlier discussion we know that the $4^{th}$ convolutional layer encodes higher-level object information related to shape and specific object parts. This indicates that object specific cues are particularly beneficial for contour detection accuracy.

We also observe that by incorporating features from all the convolutional layers, our method achieves state-of-the-art contour detection results. This suggests that the features computed by different layers are complementary and that considering information from the entire hierarchy is advantageous.

  \begin{table}
  	\begin{center}
    \begin{tabular}{ | c | c | c | c |}
    \hline
    Conv. Layers & ODS & OIS & AP\\ \hline\hline
    $1^{st}$ & 0.66 & 0.68 & 0.69 \\ \hline
	$2^{nd}$ & 0.71 & 0.74 & 0.76 \\ \hline
	$3^{rd}$ & 0.74 & 0.75 & 0.79 \\ \hline
	$4^{th}$ & 0.74 & 0.76 & 0.79\\ \hline
	$5^{th}$ & 0.73 & 0.74 & 0.77\\ \hline
	All & \bf 0.75 & \bf 0.77 & \bf 0.81\\
	
    \hline
    \end{tabular}
    \end{center}
    
        \caption{This table shows the advantage of using higher-level features from the {\em KNet} convolutional layers. Individually, the $4^{th}$ convolutional layer produces the best contour prediction results, which implies that higher-level object information is indeed beneficial for contour detection. Combining the features from all convolutional layers leads to state-of-the-art results.\vspace{-0.4cm}}
    \label{results_ind_conv}
   \end{table}

\subsection{Pooling Scheme}

When presenting the architecture of our model, we discussed three different types of pooling: {\em max}, {\em average}, and {\em center} pooling. These three techniques were used to pool the values from the sub-volumes extracted around the center point in each convolutional filter as illustrated in Fig.~\ref{fig:ss_arch}.

We now show how each type of pooling affects contour detection results. Table~\ref{results_pool} illustrates that, individually, center pooling yields the best contour detection results. This is expected because the candidate point for which we are trying to predict a contour probability is located at the center of the input patch. 

However, we note that combining all three types of pooling, achieves better contour detection results than any single pooling technique alone.

  \begin{table}
  	\begin{center}
    \begin{tabular}{ | c | c | c | c |}
    \hline
    Pooling Type& ODS & OIS & AP\\ \hline\hline
    Average & 0.73 & 0.75 & 0.78 \\ \hline
	Max & 0.69 & 0.72 & 0.73 \\ \hline
	Center & 0.74 & 0.76 & 0.8 \\ \hline
	Avg+Max+Cen & \bf 0.75 & \bf 0.77 & \bf 0.81 \\	
    \hline
    \end{tabular}
    \end{center}
    \caption{Effect of different pooling schemes on contour detection results. Center pooling produces better results than max or average pooling. Combining all three types of pooling further improves the results.\vspace{-0cm}}
    \label{results_pool}
   \end{table}

\subsection{Bifurcation and Training Objective}
\label{sec:expobj}


Next, we want to show that that the two independently-trained classification and regression branches in the bifurcated sub-network provide complementary information that yields improved contour detection accuracy. In Table~\ref{results_opt}, we present contour detection results achieved by using predictions from the individual branches of the bifurcated sub-network.

From these results, we observe that using predictions just from the classification branch produces high F-score whereas using predictions only from the regression branch yields high average precision. Combining the predictions from both branches improves the results according to both metrics thus, supporting our claim that separately optimizing edge classification and regression objectives is beneficial to contour detection.

  \begin{table}
    \begin{center}
    \begin{tabular}{ | c | c | c | c |}
    \hline
    Branch & ODS & OIS & AP\\ \hline\hline
    Classification & \bf 0.75 & 0.76 & 0.78 \\ \hline
	Regression & 0.74 & 0.76  & 0.80 \\ \hline
	Classification+Regression & \bf 0.75 & \bf 0.77 & \bf 0.81 \\	
    \hline
    \end{tabular}
    \end{center}
    \caption{Contour detection accuracy of the two branches in our bifurcated sub-network. The classification branch yields solid F-score results whereas the regression branch achieves high average precision. Averaging the outputs from these two branches further improve the results.\vspace{-0.2cm}}
    \label{results_opt}
   \end{table}

\subsection{Qualitative Results}

Finally, we present qualitative results produced by our method. In Figure~\ref{qualitative_results} we show for each input image example, the set of candidate points produced by the Canny edge detector, the un-thresholded predictions of our method, the thresholded predictions, and the ground truth contour map computed as an average of the multiple manual annotations. To generate the thresholded predictions, we use a probability threshold of $0.5$. 

Note that our method successfully distinguishes between strong and weak contours. Specifically, observe that in the last row of Figure~\ref{qualitative_results}, our method assigns lower probability to contours corresponding to zebra stripes compared to the contours of the actual object boundary separating the zebras from the background. Thus, in the thresholded version of the prediction, the weak contours inside the zebra bodies are removed and we obtain contour predictions that look very similar to the ground truth.

Due to locality of our method, it may be beneficial to apply spectral methods~\cite{Malik:2001:CTA:543015.543016,Shi97normalizedcuts} or conditional random fields~\cite{Ren_scale-invariantcontour} on top of our method to further improve its performance.

\subsection{Computational Cost}

In its current form, our method requires about $60K$ KNet evaluations ($15K$ per scale) to extract the features. Based on the runtimes reported in~\cite{jia2014caffe}, if executed on a GPU our method would take about 5 minutes and could be made faster using the approach described in~\cite{DBLP:journals/corr/IandolaMKGDK14}.

An alternative way to dramatically reduce the runtime of DeepEdge is to interpolate the entries of the feature maps produced by applying the KNet to the full image rather than to individual patches. Such an approach would reduce the number of CNN evaluations needed from $60K$ to $4$  (one for each scale), which would allow our method to run in real time even on CPUs. We note that interpolation of features in deep layers has been used successfully in several recent vision papers~\cite{DBLP:journals/corr/SermanetEZMFL13, DBLP:journals/corr/HariharanAGM14a,DBLP:journals/corr/LongSD14}. Thus, we believe that such an approach could yield nearly equivalent contour detection accuracy, up to a small possible degradation caused by interpolation.

Since in this work we were primarily interested in studying the effective advantage enabled by object-level features in contour detection, we have not invested any effort in optimizing the implementation of our method. This will be one of our immediate goals in the future.

\section{Conclusions}

In this work, we presented a multi-scale bifurcated deep network for top-down contour detection. In the past, contour detection has been approached as a bottom-up task where low-level features are engineered first, then contour detection is performed, and finally contours may be used as cues for object detection. However, due to a close relationship between object and contour detection tasks, we proposed to invert this pipeline and perform contour detection in a top-down fashion. We demonstrated how to use higher-level object cues to predict contours and showed that considering higher-level object-features leads to a substantial gain in contour detection accuracy.

Additionally, we demonstrated that our multi-scale architecture is beneficial to contour prediction as well. By considering multiple scales, our method incorporates local and global information around the candidate contour points, which leads to significantly better contour detection results. Furthermore, we showed that independent optimization of contour classification and regression objectives improves contour prediction accuracy as well. As our experiments indicate, DeepEdge achieves higher average precision results compared to any prior or concurrent work.

In conclusion, our results suggest that pure CNN systems can be applied successfully to contour detection and possibly to many other low-level vision tasks.

\section{Acknowledgements}

We thank Piotr Teterwak, Du Tran and Mohammad Haris Baig for helpful discussions. This research was funded in part by NSF award CNS-1205521.




{\small
\bibliographystyle{ieee}
\bibliography{gb_bibliography}
}

\end{document}